\documentclass{article}

\usepackage{PRIMEarxiv}

\usepackage[utf8]{inputenc} 
\usepackage[T1]{fontenc}    
\usepackage{hyperref}       
\usepackage{url}            
\usepackage{booktabs}       
\usepackage{pifont}
\usepackage{amsfonts}       
\usepackage{nicefrac}       
\usepackage{microtype}      
\usepackage{lipsum}
\usepackage{fancyhdr}       
\usepackage{graphicx}       
\graphicspath{{media/}}     
\usepackage{tabularx}
\usepackage{booktabs}
\usepackage{lscape}
\usepackage{amsmath}
\usepackage{tabularx}  
\usepackage{algorithm}
\usepackage{algorithmic}
\usepackage{multirow}
\usepackage{appendix}
\usepackage{pdflscape}
\usepackage{afterpage}
\usepackage{graphicx,amsmath,bbm,amssymb,lscape}
\usepackage{amsbsy}
\usepackage[capitalize]{cleveref}
\usepackage[table]{xcolor}
\definecolor{lightgray}{gray}{0.9}

\title{A systematic evaluation of uncertainty quantification techniques in deep learning: a case study in photoplethysmography signal analysis
\thanks{E-mails: 
Ciaran Bench (\texttt{ciaran.bench@npl.co.uk}), \ 
Oskar Pfeffer (\texttt{oskar.pfeffer@ptb.de}), \ 
Vivek Desai (\texttt{vivek.desai@npl.co.uk}), \ 
Mohammad Moulaeifard (\texttt{mohammad.moulaeifard@uol.de}), \ 
Lo\"ic Coquelin (\texttt{loic.coquelin@lne.fr}), \ 
Peter H. Charlton (\texttt{pc657@cam.ac.uk}), \ 
Nils Strodthoff (\texttt{nils.strodthoff@uol.de}), \ 
Nando Hegemann (\texttt{nando.hegemann@ptb.de}), \
Philip J. Aston (\texttt{philip.aston@npl.co.uk}), \ 
Andrew Thompson (\texttt{andrew.thompson@npl.co.uk})
}
}

\begin{document}
\maketitle

Ciaran Bench,$^1$  
Oskar Pfeffer,$^2$  
Vivek Desai,$^1$ 
Mohammad Moulaeifard,$^3$ 
Lo\"ic Coquelin,$^4$
Peter H. Charlton,$^5$ \\
Nils Strodthoff,$^3$ 
Nando Hegemann,$^2$
Philip J. Aston,$^{1,6}$ 
Andrew Thompson$^1$ \\[0.15cm]
$^1$Department of Data Science and AI, National Physical Laboratory, Teddington, United Kingdom \\
$^2$Physikalisch-Technische Bundesanstalt, Berlin, Germany \\
$^3$Carl von Ossietzky Universit\"at Oldenburg, Oldenburg, Germany \\
$^4$Laboratoire National de M\'etrologie et d'Essais, Paris, France \\
$^5$Department of Public Health and Primary Care, University of Cambridge, Cambridge, United Kingdom \\
$^6$School of Mathematics and Physics, University of Surrey, Guildford, United Kingdom
\vspace{.2in}

\begin{abstract}
In principle, deep learning models trained on medical time-series, including wearable photoplethysmography (PPG) sensor data, can provide a means to continuously monitor physiological parameters outside of clinical settings. However, there is considerable risk of poor performance when deployed in practical measurement scenarios leading to negative patient outcomes. Reliable uncertainties accompanying predictions can provide guidance to clinicians in their interpretation of the trustworthiness of model outputs. It is therefore of interest to compare the effectiveness of different approaches. Here we implement an unprecedented set of eight uncertainty quantification (UQ) techniques to models trained on two clinically relevant prediction tasks: Atrial Fibrillation (AF) detection (classification), and two variants of blood pressure regression. We formulate a comprehensive evaluation procedure to enable a rigorous comparison of these approaches. We observe a complex picture of uncertainty reliability across the different techniques, where the most optimal for a given task depends on the chosen expression of uncertainty, evaluation metric, and scale of reliability assessed. We find that assessing local calibration and adaptivity provides practically relevant insights about model behaviour that otherwise cannot be acquired using more commonly implemented global reliability metrics. We emphasise that criteria for evaluating UQ techniques should cater to the model's practical use case, where the use of a small number of measurements per patient places a premium on achieving small-scale reliability for the chosen expression of uncertainty, while preserving as much predictive performance as possible. 

\end{abstract}


\section{Introduction}

While there is considerable precedent demonstrating that deep learning models can achieve state of the art performance on a wide range of predictive tasks, there remains a significant risk of poor performance when deployed in more practical measurement scenarios. This (among other reasons) has hindered their routine use in domains with stringent requirements for accuracy, such as medical sensing. 

Some information about the underlying doubt in a given prediction can establish its trustworthiness, facilitating the effective use of the model in more realistic settings (e.g.\ enabling one to disregard predictions likely to be incorrect). Consequently, there is significant interest in both formulating effective methods to assess the trustworthiness of predictions, and in developing evaluation metrics to determine the quality of this assessment of trustworthiness. This can help determine the most useful approaches for a given task.

\begin{figure}[h]
    \centering
    \includegraphics[width=0.8\linewidth]{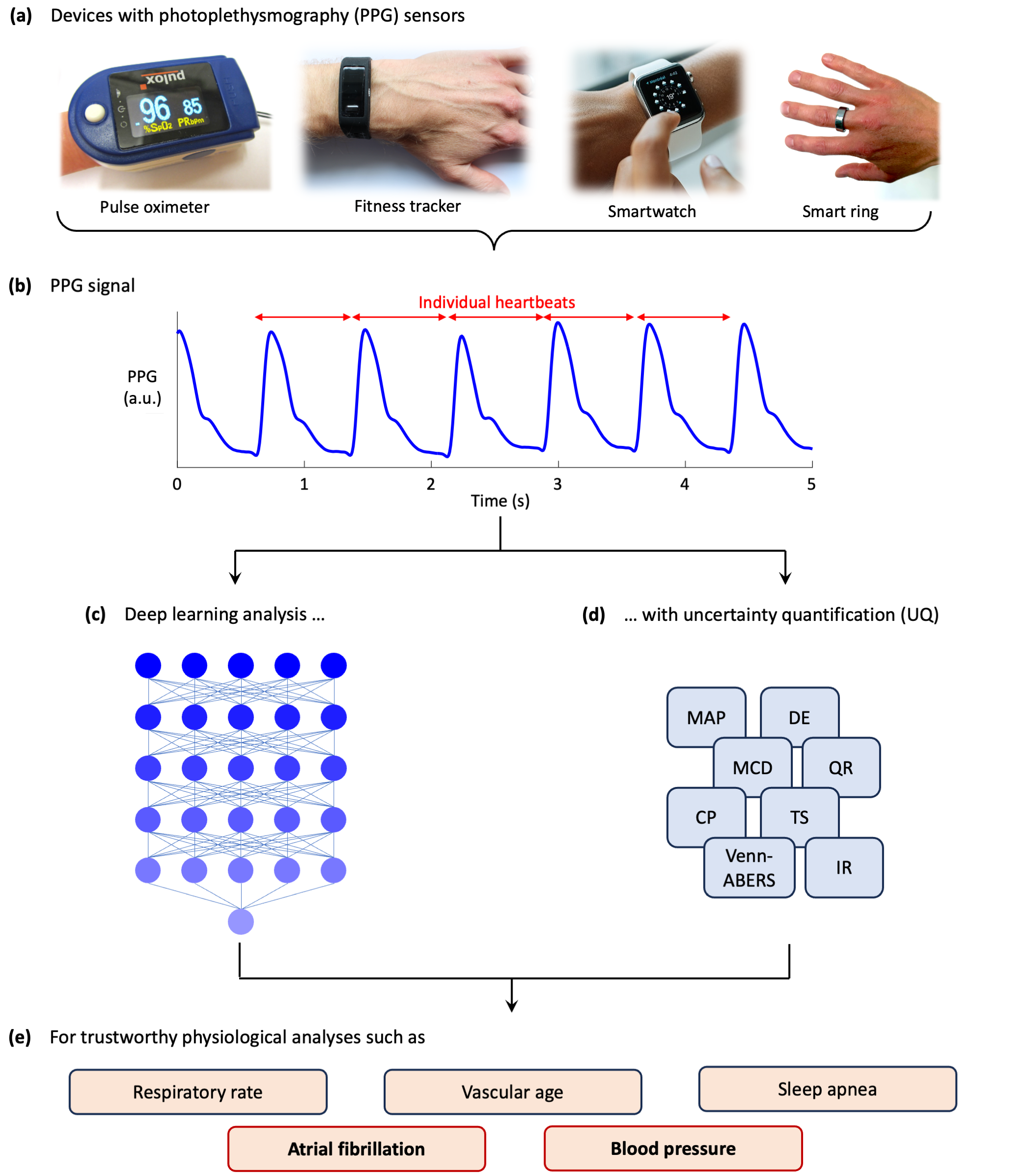}
    
    \caption{Uncertainty quantification in deep learning analyses of photoplethysmography (PPG) signals: (a) PPG signals can be measured by many clinical and consumer devices; (b) PPG signals capture the pulsation of blood with each heartbeat; (c) deep learning is commonly used to analyse PPG signals; (d) this study provides a systematic evaluation of uncertainty quantification techniques for deep learning; (e) aiming to improve the trustworthiness of analyses.}
    \label{fig:uq_for_ppg}
\end{figure}

Here, we aim to compare the effectiveness of several uncertainty quantification (UQ) techniques, each of which aims to provide an explicit estimate of the degree of doubt in a model's prediction. To enable a thorough evaluation, we formulate a comprehensive evaluation framework for assessing the reliability of the predicted uncertainties. We apply the UQ techniques and our uncertainty evaluation framework to a type of data which is typical of many potential use cases requiring UQ: the analysis of photoplethysmography (PPG) signals - physiological signals which are widely measured by clinical and consumer devices, and used to inform clinical decision making. This concept is illustrated in Figure~\ref{fig:uq_for_ppg}.

\subsection{Case study: Photoplethysmography (PPG)}

PPG is an optical sensing technique which consists of shining light on to a bed of tissue (such as the finger or wrist), and measuring the amount of light either transmitted through or reflected from the tissue over time. PPG is widely used to measure the pulse - the pulsation of blood which occurs with each heartbeat (see Figure~\ref{fig:uq_for_ppg}(b)). PPG came into clinical use in the 1980s in the form of pulse oximeters, non-invasive devices which clip on the finger and provide non-invasive measurements of blood oxygen saturation and heart rate \cite{Miyasaka2021}. Pulse oximeters remain widely used across many healthcare settings, from hospitals to the home. More recently, the technology has also been incorporated into wearables such as smartwatches, smart rings, and fitness trackers, where it is used for unobtrusive physiological monitoring \cite{Charlton2021d}. Typical PPG-based devices are illustrated in Figure~\ref{fig:uq_for_ppg}(a). 

The analysis of PPG signals is a typical use case for UQ: deep learning models for PPG signals are subject to inaccuracies due to the complexity of the signal and the underlying physiology; and PPG analyses need to be trustworthy as they are used to inform clinical decision making. Indeed, parameters derived from PPG signals are being used increasingly in healthcare, including not only the traditional use of PPG in pulse oximeters for oxygen saturation monitoring, but also emerging applications for wearable-based PPG measurements, such as detecting atrial fibrillation \cite{Perez2019} and cuffless blood pressure monitoring \cite{karioInnovations2024a}. These applications are potentially highly beneficial as they could enable the detection of undiagnosed disease, and continuous monitoring of physiology which could otherwise only be monitored by a specialist device. For instance, atrial fibrillation and hypertension (elevated blood pressure) are often not diagnosed \cite{turakhiaContemporary2023a,officefornationalstatisticsonsRisk2023}. In addition, recent developments could enable continuous monitoring of atrial fibrillation \cite{zhaoEvaluation2024} and blood pressure \cite{karioInnovations2024a}, providing richer information than can be acquired in a single clinic visit which could inform treatment decisions such as in the management of hypertension \cite{krakoff2016blood,o2008ambulatory,piper2015diagnostic,parati2008european}.

While several classical signal processing approaches have been used to optimise predictive models for a range of tasks (e.g.\ assessing arterial stiffness \cite{Charlton2021} and estimating respiratory rate \cite{Charlton2016} from the PPG), deep learning models provide a convenient means to handle the large quantities of data captured by wearable devices, and a generic framework for learning the optimal nonlinear transformation  of the data without needing to rely on potentially inaccurate prior knowledge about its analytical form. Similarly, their capacity for learned/automatic feature extraction avoids limitations with using prior knowledge to formulate hand-crafted features.
Models have been trained to predict blood pressure from PPG time series data \cite{gonzalez2023benchmark,el2020review,maqsood2022survey,paviglianiti2022comparison}, helping to overcome challenges with the utilisation of multiple sensor measurements (e.g.\  mismatched sampling rates, synchronisation difficulties) needed for pulse-transit time (PTT) based measurements, and offering a potential means to side-step the need for laborious per-patient calibration also associated with PTT \cite{el2020review}. Comparable efforts have been made for AF classification \cite{shashikumar2017deep,tison2018passive,shen2019ambulatory,gotlibovych2018end}. 

In this study, we focus on two prediction tasks: atrial fibrillation detection and blood pressure estimation. For both prediction tasks, we train our models with fixed-length, raw (minimal preprocessing), 1D PPG time series as an input. For BP prediction, the model simultaneously predicts the patient's systolic blood pressure (SBP) and diastolic blood pressure (DBP) at the time of measurement. For AF, the model predicts whether the patient has endured an episode of AF within the duration of the measurement (binary classification).

\subsection{The need for uncertainty quantification and the evaluation of uncertainty reliability in PPG analysis}
Models implemented in realistic settings will routinely encounter unseen examples from patients with a diverse range of characteristics and routines, leaving considerable risk for misdiagnosis. Therefore, in addition to achieving satisfactory predictive accuracy, some notion of the uncertainty in the models' outputs is needed to establish whether they can be used to reliably inform diagnosis. It is of interest to determine which UQ techniques may be most suitable, necessitating a comprehensive evaluation framework for uncertainty reliability. Yet, the existing literature related to medical time series is unsatisfactory, either considering only a small subset of UQ techniques \cite{song2023uncertainty,vranken2021uncertainty}, uncertainty reliability metrics, or smaller datasets that do not reflect the full variation in signal properties/joint relationship between input feature and the ground truth that may be encountered in realistic measurement settings \cite{song2023uncertainty,harper2019end,trudaquantifying,liu2022videocad,asgharnezhad2023improving,han2023non,das2020bayesbeat,belhasin2024uncertainty,vranken2021uncertainty,chen2022quantifying}.

\subsection{Our contributions}
Here, we compare an unprecedented set of eight UQ techniques applied to deep learning models trained on two physiologically relevant prediction tasks: Atrial Fibrillation (AF) classification and two variants of a blood pressure (BP) regression task.  Our choice of UQ techniques enables a comparative study of how various theoretical frameworks for uncertainty quantification (e.g.\ frequentist, Bayesian, and heuristic ensembles), sources of uncertainty considered (epistemic/aleatoric), and quantification strategies (e.g.\ post-hoc recalibration, intrinsic modelling, and ensembling) affect reliability. We broadly categorise our approaches into three classes: post-hoc recalibration, post-hoc ensemble, and intrinsic techniques. These are summarised in Table \ref{tab:UQ_types_descriptions}. 

We also formulate a comprehensive evaluation framework (assessing both predictive performance and the quality of uncertainties i.e.\ their reliability), and use large-scale and realistic datasets to enable a thorough comparison of the different UQ approaches under realistic measurement conditions. We highlight the practical challenges with comparing UQ techniques that produce different output types, and suggest conversion schemes to enable a comparison of uncertainty reliability across a broad range of UQ approaches.

\begin{table}[h!]
\centering
\small
\caption{Uncertainty quantification methodology and descriptions}
\begin{tabular}{|l|l|p{0.35\textwidth}|p{0.35\textwidth}|}
\hline
\multicolumn{2}{|c|}{\textbf{UQ Methodology}} & \textbf{Description} & \textbf{UQ Techniques} \\
\hline
\multicolumn{2}{|l|}{Intrinsic} & Uncertainties are estimated as a consequence of the design choices involved with the model's optimisation/evaluation.
& Maximum \textit{a Posteriori} (MAP) Estimation, Monte-Carlo Dropout, Quantile Regression \\
\hline

\hline
\multirow{2}{*}{Post-hoc} & Post-hoc ensemble & Multiple versions of the same architecture are trained to perform the same task, where uncertainties are derived by aggregating their outputs. Can be applied to multiple instances of an intrinsic UQ technique, and their respective post-hoc recalibrations if available. & Deep Ensembles \\
\cline{2-4}
& Post-hoc recalibration & A learned transformation (typically optimised over a held-out calibration set) is applied to the predictions to improve their calibration. Can be applied to the outputs of intrinsic, or post-hoc ensemble UQ techniques & Temperature Scaling, Conformal Prediction, Isotonic Regression \\
\hline

\end{tabular}
\label{tab:UQ_types_descriptions}
\end{table}

\begin{table}[h!]
\footnotesize
\centering
\caption{Uncertainty quantification techniques. (C) indicates classification, and (R) represents regression. * Frequentist techniques do not explicitly model different sources of uncertainty. Post-hoc techniques instead aim to improve the reliability of informative, but poorly calibrated, uncertainties outputted by a model. Therefore, the sources of uncertainty encoded in the outputs depend on those captured by the base model/UQ technique. MAP and QR have output types that straightforwardly express uncertainty, and while we express this uncertainty here, this is not always implemented in practical use cases. 
}
\begin{tabularx}{\textwidth}{|X|X|c|X|X|X|}
\hline
\textbf{UQ Technique} & \textbf{Theoretical Formulation} & \textbf{Reg/Class} & \textbf{Epistemic/Aleatoric} & \textbf{Expressions} & \textbf{Output Type}\\
\hline
MAP Estimation (MAP) & Bayesian & \checkmark/\checkmark & \ding{55} (point estimate)/\checkmark & R: Predicted variance, C: Class probabilities or their entropy & R: mean and variance parametrising Gaussian, C: binary class distribution \\
\hline
Monte Carlo Dropout (MCD) & Bayesian Inspired/Approximate Variational Learning & \checkmark/\checkmark & \checkmark/\checkmark (with likelihood-based loss) & R: Law of Total Variance - sum of mean of predictive variances, and variance of predicted means, C: Aggregated noise-corrupted class probabilities or their entropy & R: mean and variance parametrising Gaussian, C: binary class distribution and logit variances\\
\hline
Quantile Regression (QR) & Frequentist & \checkmark/\ding{55} & *  & Variance of Gaussian parametrised by predicted quantiles & Quantiles \\
\hline
Deep Ensembles (DE) & Heuristic ensemble (has Bayesian interpretation) & \checkmark/\checkmark  & \checkmark/\checkmark (with likelihood-based loss) & R: Law of Total Variance - sum of mean of predictive variances, and variance of predicted means C: Aggregated noise-corrupted class probabilities or their entropy & R: mean and variance parametrising Gaussian, C: binary class distribution and logit variances\\
\hline
Conformal Prediction (CP) & Frequentist & \checkmark/\ding{55} & * & Same as `uncalibrated' uncertainties (here, for regression, it's the variance of a Gaussian parametrised by prediction intervals) & Prediction intervals for the base model output type with coverage guarantees\\
\hline
Venn-ABERS Conformal Prediction (Venn-ABERS) & Frequentist & \ding{55}/\checkmark & 
* & Class probabilities 
& Probability intervals for a given class\\
\hline
Temperature Scaling (TS) & Frequentist & \checkmark/\checkmark & * & Same as `uncalibrated' uncertainties & Base model output type\\
\hline
Isotonic Regression (IR) & Frequentist & \checkmark/\checkmark & * & Same as `uncalibrated' uncertainties & Base model output type \\
\hline

\end{tabularx}
\label{tab:UQ_techniques}
\end{table}

\begin{table}[h]
\centering
\caption{Uncertainty Quantification Techniques and Metrics}
\begin{tabular}{ll|ll}
\toprule
\multicolumn{2}{c|}{\textbf{UQ Techniques}} & \multicolumn{2}{c}{\textbf{Metrics}} \\
\midrule
MAP & Maximum \textit{a posteriori} & \multicolumn{2}{l}{\textbf{Classification}} \\ 
MCD & Monte Carlo Dropout & ECE & Expected Calibration Error \\
DE & Deep Ensembles & smECE & Smooth Expected Calibration Error \\
QR & Quantile Regression & ACE & Adaptive Calibration Error \\
CP & Conformal Prediction & VCE & Variation Calibration Error \\
Venn-ABERS & Venn-ABERS Conformal Prediction & UCE & Uncertainty Calibration Error \\
TS & Temperature Scaling & AUC & Area Under the (ROC) Curve \\
IR & Isotonic Regression & & \\
\cmidrule(lr){1-4}
 &  & \multicolumn{2}{l}{\textbf{Regression}} \\ 
 &  & ENCE & Expected Normalised Calibration Error \\
 &  & CCE & Coverage Calibration Error \\
 &  & CRPS & Continuous Ranked Probability Score \\
 &  & PICP & Prediction Interval Coverage Probability \\
 &  & NLL & Negative Log-Likelihood \\
 &  & MAE & Mean Absolute Error \\
 &  & MASE & Mean Absolute Scaled Error \\
\cmidrule(lr){3-4}
\end{tabular}
\label{tab:abb}
\end{table}

\section{Methods}
\subsection{UQ in deep learning}

Classical UQ approaches, such as Bayesian inference, do not scale well with the number of parameters involved with deep networks. This has spurred the development of more scalable UQ approaches, each of which may employ different frameworks for uncertainty quantification, model various sources of uncertainty, and may have different output types, leading to varied expressions of uncertainty.  We implement a broad range of popular and scalable UQ techniques, including: Monte Carlo Dropout (MCD) , Deep Ensembles (DE), and different variants of Conformal Prediction (CP) and post-hoc recalibration techniques. 

We also include Quantile Regression (QR) and Maximum \textit{a posteriori} estimation (MAP), whose outputs encode an interpretable uncertainty, but are not typically expressed in practical scenarios. Each UQ technique is briefly described in Table \ref{tab:UQ_techniques}, while Appendix \ref{sec:UQ_methods} contains detailed information about the theory underpinning each technique, as well as implementation details. These techniques may be categorised into three quantification strategies described in Table \ref{tab:UQ_types_descriptions}: intrinsic modelling, and post-hoc techniques (post-hoc recalibration, and post-hoc ensemble). The abbreviations for the UQ methods and evaluation metrics are given in Table \ref{tab:abb}.

Ultimately, a reliable uncertainty should provide an accurate indication of the degree of doubt in a given prediction. The theory underpinning each technique is likely to impact reliability. In terms of the UQ frameworks, the frequentist approaches consider uncertainty as a long-run frequency of occurrence, but may not provide accurate estimates when data is scarce (a common reality in medical applications). In contrast, the Bayesian framework uses prior beliefs updated with evidence to derive predictive distributions. However, ill-defined priors and other computational challenges may result in unreliable uncertainties. Deep ensembles is distinct in that it employs a heuristic ensemble framework, where a discrete predictive distribution is acquired by collating the outputs produced by a set of identical but independently trained models (each parametrised with a unique random weight initialisation). However, it is resource intensive making it challenging to optimise its parametrisation (e.g.\ the number of models in the ensemble).

There are various sources of uncertainty which contribute to the overall predictive uncertainty. Understanding the impacts of these different sources can help to provide the most accurate predicted uncertainty, and also to inform decisions pertaining to dataset curation, the choice of model architecture, and optimisation strategy \cite{kendall2017uncertainties}. Two types of uncertainty in neural networks are often distinguished: \emph{aleatoric} and \emph{epistemic} uncertainty. Epistemic uncertainty is typically thought of as `model' uncertainty, and may be reduced by training the model on more or higher quality data, or by choosing a model architecture that is better equipped to detect task-specific features or has greater capacity~\cite{valdenegro2022deeper}. Aleatoric uncertainty refers to irreducible uncertainty, which could arise either from uncertainty in the data fed into the model or from ill-posedness \cite{Goodfellow2016DeepLearning} of the learning task. Accurate estimates of aleatoric uncertainty could help inform data collection protocols and provide insights into the feasibility of prediction tasks. Here, we hope to observe the extent to which all of these factors may affect uncertainty reliability. 

\subsubsection{UQ methods}
For regression we implement MAP, MCD, DE, and QR, along with the subsequent application of post-hoc recalibration techniques to QR and DE. We present results for IR, TS for both DE and QR techniques. Conformal prediction is applied to MAP and QR. The full implementation details for conformal prediction (e.g.\ different score functions) can be found in Section~\ref{appendix:conformal} in the Appendix. The recalibration methods are implemented using the \textsc{netcal} Python package for regression \cite{Kueppers_2022_ECCV_Workshops}.
For the conformal method, and for QR, results are given as confidence intervals corresponding to  $1$ and $2$ Gaussian standard deviations from the mean, namely the quantile intervals $[0.1587,0.8413]$ and $[0.0228,0.9772]$ respectively; these confidence intervals are denoted in the results as $1\sigma$ and $2\sigma$ respectively. For classification, we implement MAP, MCD, DE, Venn-ABERS, and finally post-hoc recalibration (TS and IR) applied to MCD and DE.

\subsection{Evaluating uncertainty reliability}

To realise the practical benefits of using estimated uncertainties, it is essential that models not only provide estimates, but that they are also reliable, reflecting the true underlying doubt in a given prediction \cite{pernot2023calibration}. Calibration is a related concept often referenced in studies on uncertainty quantification (UQ). This refers to the extent to which the predicted uncertainties align with the true frequency of outcomes, often assessed by comparing the magnitude of uncertainties with prediction error. It is central to some of the UQ techniques implemented in this work (e.g.\ post-hoc recalibration techniques). However, as will be discussed (Appendix \ref{sec:uq_eval}), there are other notions of reliability (e.g.\ sharpness) that are relevant to assessing the trustworthiness of model predictions. We make this distinction where necessary.

The quality of uncertainty estimates must be assessed carefully, and should cater to the practical realities of a model's clinical use case; in particular, that just a single measurement, or a few measurements, will often be used to inform diagnosis. In the ideal case, an estimate of uncertainty accompanying a single predicted quantity should reflect the underlying doubt in this single prediction (i.e.\ the model must exhibit `individual reliability', a term that will be discussed in Appendix \ref{sec:local_global}). With that said, most uncertainty validation metrics evaluate reliability over populations of estimates, raising complications with formulating an appropriate and effective evaluation framework \cite{pernot2023calibration}. Nevertheless, the consequences are clear: unreliable uncertainties will suggest incorrect degrees of doubt in a given prediction, increasing the potential risk of misdiagnosis and negative patient outcomes.

Here, we implement a range of reliability metrics and visualisations that indicate:

\begin{itemize}
    \item Global reliability (reliability assessed over the whole test set)
    \item Local/small scale reliability (reliability assessed per bin of examples, binned by magnitude) \cite{pernot2023calibration}
    \item Adaptivity (reliability assessed per bin of examples, binned by something other than magnitude) \cite{pernot2023calibration,bench2025uncertainty}
    \item Sharpness (extent to which confidence intervals concentrate around the predicted value)~\cite{gneiting2007probabilistic}
    \item Calibration (extent to which the estimated uncertainties correlate with prediction error).
\end{itemize}

We also include proper scoring rules (metrics that are optimal when the predictive distribution matches the true distribution; i.e.\ considering both calibration and sharpness), metrics that employ various binning strategies, and cater to various expressions of uncertainty. We also consider metrics that employ the variance and coverage based frameworks for assessing reliability (the former assessing whether the predicted uncertainty is equivalent to prediction error, while the latter considers whether the frequency with which the ground truth occurs within a given confidence interval of the predicted distribution equals the confidence), and employ conversions between different output types to enable a comprehensive comparison of UQ techniques with distinct output types. 

Appendix \ref{sec:uq_eval} provides an overview of the principles underpinning the uncertainty evaluation framework.  Tables \ref{tab:uncertainty_metrics} and \ref{tab:uncertainty_metrics_reg} provide a summary of the metrics, and Appendix \ref{sec:appendix_eval_metrics} provides more in depth  descriptions of each metric. We also provide predictive accuracy metrics alongside uncertainty reliability metrics. For AF classification, we give the Area Under the (Receiver Operating Characteristic (ROC)) Curve (AUC), which determines the probability that a classifier will rank a randomly selected positive instance above a randomly selected negative instance \cite{hanley1982meaning}.

For regression, we provide the mean absolute prediction error (MAE), and mean absolute scaled prediction error (MASE, which is the MAE divided by the prediction error from using the training set median \cite{moulaeifard2025machine}) for SBP and DBP. We also provide an additional visualisation of small-scale calibration for regression tasks by plotting a bivariate histogram of prediction error against predicted uncertainty.

\begin{table}[htbp]
\footnotesize
\caption{Uncertainty reliability metrics for classification.}
\label{tab:uncertainty_metrics}
\begin{tabularx}{\textwidth}{|X|X|c|X|X|X|}
\hline
\textbf{Metric} & \textbf{Expression} & \textbf{Locality} & \textbf{Visualisation} & \textbf{Proper scoring rule} & \textbf{Description} \\
\hline
Expected Calibration Error (ECE)~\cite{guo2017calibration} & Predicted class probability & Local & Average confidence in bin vs.\ average accuracy & \ding{55} & Weighted average of the squared difference between the average magnitude of the predicted class probabilities and the corresponding mean accuracy of predictions binned by magnitude of the predicted class probabilities. \\
\hline
Smooth Expected Calibration Error (smECE)~\cite{blasiok2023smooth} & Predicted class probability & Local & Kernel smoothed confidence vs accuracy & \ding{55} & ECE variant that uses kernel density estimation; mitigates binning related drawbacks of ECE.\\
\hline
Adaptive Calibration Error (ACE)~\cite{nixon2019measuring} & Predicted class probability & Local & Average confidence in bin vs.\ accuracy, with adaptive binning & \ding{55} & ECE variant with adaptive binning, accommodating sparse predictions and class imbalance.\\
\hline
Variation Calibration Error (VCE)~\cite{thompson2025variation} & Compatible with several expressions of variation (here we use entropy) & Local & Mean predicted uncertainty (expressed here as entropy) vs.\ mean observed variation (also expressed as an entropy) in each uncertainty bin & \ding{55} & Weighted average of the squared difference between the  average entropy of the predicted distributions and the corresponding observed variation binned by magnitude of the predicted entropies.\\
\hline
Uncertainty Calibration Error (UCE)~\cite{lavesuncertainty} & Entropy of predicted probability distribution & Local & Mean predicted uncertainty (expressed as entropy) vs.\ mean observed error in each uncertainty bin & \ding{55} & Weighted average of the squared difference between the  average entropy of the predicted distributions and the corresponding mean inaccuracy of predictions binned by magnitude of the entropies.\\
\hline
Negative Log Likelihood (NLL) & Predicted class distribution & Global & N/A & \checkmark & Proper scoring rule; the sum of the log probabilities of the target classes.\\
\hline
\end{tabularx}
\end{table}

\begin{table}[htbp]
\caption{Uncertainty reliability metrics for regression. }
\footnotesize
\label{tab:uncertainty_metrics_reg}
\begin{tabularx}{\textwidth}{|X|X|c|X|X|c|X|}
\hline
\textbf{Metric} & \textbf{Input} & \textbf{Locality} & \textbf{Visualisation} & \textbf{Proper Scoring Rule} & \textbf{Var/Coverage}& \textbf{Description}\\
\hline
Expected Normalised Calibration Error (ENCE)~\cite{levi2022evaluating} & Predictive distribution & Local & Root mean square of the prediction errors in a bin vs.\ the root mean of the corresponding predicted variances & \ding{55} & V & Weighted average of the absolute difference between the root mean square prediction error, and the corresponding root mean of the predicted variance for each uncertainty bin.\\
\hline
Continuous Ranked Probability Score (CRPS)~\cite{grimit2006continuous} & Predictive distributions or quantiles & Global & N/A & \checkmark & C & The integrated squared difference between the empirically observed CDF (step function) and that acquired from the predictive distribution.\\
\hline
Prediction Interval Coverage Probability (PICP) ~~\cite{kuleshov2018accurate} & Prediction intervals & Global & N/A & \ding{55} & C & The ratio of observations that fall within the interval defined by a chosen coverage level (we choose 1$\sigma$ and 2$\sigma$ of the assumed form of the predicted distribution) divided by the chosen coverage level. Optimal value is 1.\\
\hline
Coverage Calibration Error (CCE) ~\cite{kuleshov2018accurate}& Prediction intervals & Global & Plot of coverage level vs.\ observed coverage for various coverage levels & \ding{55} & C & The average of the squared differences between the chosen coverage level and the ratio of observations that fall within the interval defined by this coverage level, over various coverage levels.\\
\hline
\end{tabularx}
\end{table}

\subsection{Data}
\subsubsection{BP estimation task}
        
For the BP estimation regression task, the VitalDB dataset is used, which includes ECG, PPG, and invasive arterial blood pressure (ABP) signals from surgical patients \cite{lee2022vitaldb}. Wang et al. \cite{wang2023pulsedb} released a pre-processed version as part of the PulseDB dataset. From this, we extracted 10-second PPG segments (125 Hz sampling frequency) along with reference systolic and diastolic blood pressures derived from ABP signals (Table \ref{table_vitalDB}). The dataset supports both calibration-based (distinct from uncertainty calibration, and instead refers to having data from the same patients being included in the various splits) and calibration-free (distinct subjects, i.e.\ no overlap in patient data across the dataset splits) testing—referred to here as VitalDB ‘calib’ and ‘calibfree’ respectively—which is vital for evaluating the generalisability of BP estimation models. To align with previous studies \cite{moulaeifard2025machine}, we retained the original test sets and partitioned the training sets into training, validation, and calibration subsets, reflecting the original test set construction. 

\begin{table}[ht]
\centering
\caption{Characteristics of the VitalDB subsets used for BP estimation. The table is taken from \cite{moulaeifard2025machine}.}
\label{table_vitalDB}
\begin{tabular}{|p{0.35\columnwidth}|p{0.28\columnwidth}|p{0.28\columnwidth}|}
\hline
Subset & VitalDB `Calib' & VitalDB `CalibFree' \\ \hline
Train (samples / subjects)       & 418986 / 1293   & 416880 / 1158   \\ \hline
Validation (samples / subjects)  & 40673 / 1293    & 32400 / 90      \\ \hline
Calibration (samples / subjects)  & 40673 / 1293    & 16200 / 45      \\ \hline
Test (samples / subjects)        & 51720 / 1293    & 57600 / 144     \\ \hline
Age (years, mean $\pm$ SD)              & 58.98 $\pm$ 15.03 & 58.89 $\pm$ 15.07 \\ \hline
Sex (M\%)                       & 57.69           & 57.91           \\ \hline
SBP (mmHg, mean $\pm$ SD)        & 115.48 $\pm$ 18.92 & 115.47 $\pm$ 18.91 \\ \hline
DBP (mmHg, mean $\pm$ SD)        & 62.92 $\pm$ 12.08  & 62.93 $\pm$ 12.06  \\ \hline
\end{tabular}
\end{table}

\subsubsection{AF detection task}

For the AF classification task, we used the DeepBeat dataset \cite{torres2020multi}. It includes over 500,000 25-second PPG segments sampled at 32 Hz from 175 individuals (108 with AF, 67 without), recorded via a wrist-worn device from participants before cardioversion, during exercise stress tests, and in daily life. The original split resulted in overestimated performance metrics due to an imbalanced distribution of AF and non-AF cases. To address this, we rely on a new, subject-level split that eliminates overlap and balances the AF/non-AF ratio across training, validation, calibration and test sets (Table \ref{table_AF}), based on the ratio proposed in \cite{moulaeifard2025machine}.

\begin{table}[ht]
\centering
\caption{Characteristics of the DeepBeat subsets used for AF classification.The table is taken from \cite{moulaeifard2025machine}.}
\label{table_AF}
\begin{tabular}{|p{0.2\columnwidth}|p{0.2\columnwidth}|p{0.15\columnwidth}|p{0.12\columnwidth}|p{0.12\columnwidth}|}
\hline
Dataset & \multicolumn{4}{c|}{DeepBeat (AF classification)} \\ \hline
Subset  & AF                      & Non-AF               & Data Ratio & AF Ratio \\ \hline
Train (samples / subjects)      & 40603 / 50    & 65646 / 38    & 0.70      & 0.38   \\ \hline
Validation (samples / subjects) & 5800 / 19     & 9456 / 7      & 0.10      & 0.38   \\ \hline
Calibration (samples / subjects) & 5808 / 20     & 9273 / 14      & 0.10      & 0.38   \\ \hline
Test (samples / subjects)       & 5797 / 19     & 9580 / 5      & 0.10      & 0.37   \\ \hline
\end{tabular}
\end{table}

\subsection{Models}
Both the predictive performance for AF and BP prediction models \cite{moulaeifard2025machine} and reliability for a given UQ technique can vary considerably depending on the chosen architecture \cite{he2023survey}. We compare the performance of two models: a larger capacity, residual block-based xresnet1d50 (referred to here as \textbf{resnet}), and a 1D-variant of the more generic AlexNet (referred to as \textbf{alexnet}). The training procedures used for each UQ technique are given in Appendix \ref{sec:train_details}. The xresnet1d50 model is a 50-block 1D variant of the xresnet architecture, that offers several improvements to the generic resnet that help increase predictive accuracy \cite{he2019bag,strodthoff2023ptb}. The 1D-variant has been show to produce more accurate predictions on a range of ECG \cite{strodthoff2023ptb} and PPG-based classification and regression tasks compared to other model architectures \cite{bench2024towards}, even on the same PPG regression datasets used in these reports \cite{moulaeifard2025machine, moulaeifard2025generalizable}. Similarly, the AlexNet1D model has been shown to produce highly accurate predictions relative to other models for some PPG predictions tasks \cite{moulaeifard2025machine}. 

Here, we use MAP estimation as a baseline for both tasks. For regression, we employ a Gaussian negative log-likelihood loss (GNLL). All models predict both SBP and DBP simultaneously.  AF classification is posed as a binary classification task, where a custom loss for modelling the logit variance is implemented and used for MAP estimation as the baseline.

\section{Results and discussion}
\subsection{AF classification}
\subsubsection{Global evaluation}\label{sec:global}
Global uncertainty evaluation results for AF classification are shown in Tables~\ref{table:AF alexnet} and~\ref{table:AF resnet}.  Table~\ref{table:AF alexnet} shows the results for the \textbf{alexnet} model and Table~\ref{table:AF resnet} shows the results for the \textbf{resnet} model. An expanded set of predictive performance metrics are given in Appendix \ref{sec:appendix_pred_performance}.

It is evident from these tables that the optimal UQ method depends on the expression of uncertainty, and the chosen evaluation metric. ECE, ACE, smECE and NLL cater to class probabilities/confidence as the expression of uncertainty, and UQ methods which optimise reliability for this expression of uncertainty have correspondingly low values. In contrast, these methods exhibit poor (high) UCE values and instead \textbf{alexnet} DE produced the best UCE of 0.043. We believe that the main reason for this difference is that in the UCE predicted entropies are not being compared `like-for-like' with averaged observed entropies but instead with observed misclassification proportions; see Appendix~\ref{sec:UCE} for further details. The results for VCE, on the other hand, tell a similar story to the other calibration metrics.

The Venn-ABERS and post-hoc recalibration techniques improve the reliability of the predicted confidences considerably for both models. In particular, Isotonic Regression (IR) gave the best confidence calibration results, with the \textbf{alexnet} MCD+IR method obtaining an ECE and smECE of 0.048 and 0.044 respectively. This is in line with their expected behaviour, given they are developed to improve the calibration of predicted confidences. Indeed, one would not expect to see improvements in metrics that do not align with the optimisation target of the UQ post-hoc technique. For a different optimisation objective one should consider a different calibration/conformal prediction approach that is tailored to this specific target.

For both models, the post-hoc recalibration and Venn-ABERS methods obtain superior performance with respect to the NLL metric which, being a proper scoring rule, captures both calibration and sharpness. Across both models, the MAP estimation method produces the least reliable uncertainties according to several of the metrics. 

Despite various differences in formulation/properties of the confidence-based reliability metrics implemented here, none produced noteworthy differences in the ranking of the various UQ techniques. This suggests that differences in binning strategy between the ECE and ACE, as well as the smoothing of binning with the smECE, and the use of entropy as a measure of variation in VCE, have a relatively minor impact on the effectiveness of the metrics.

\begin{figure}[h]
    \centering
    \includegraphics[width=0.4\linewidth]{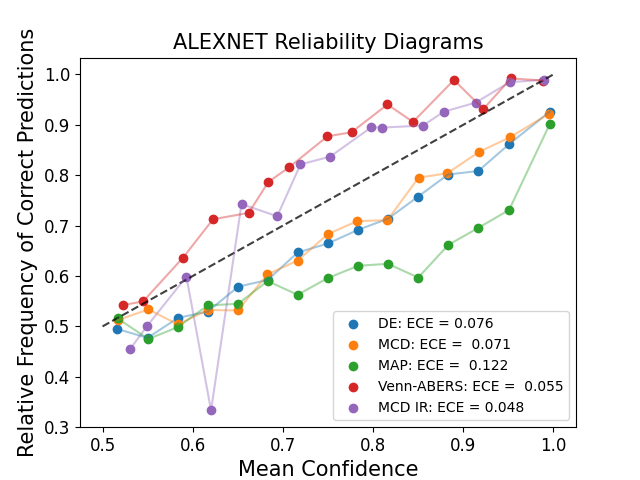}
    \includegraphics[width=0.4\linewidth]{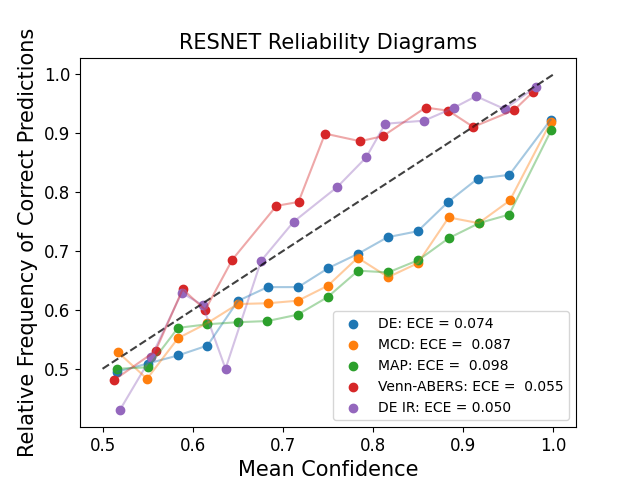}
    \includegraphics[width=0.4\linewidth]{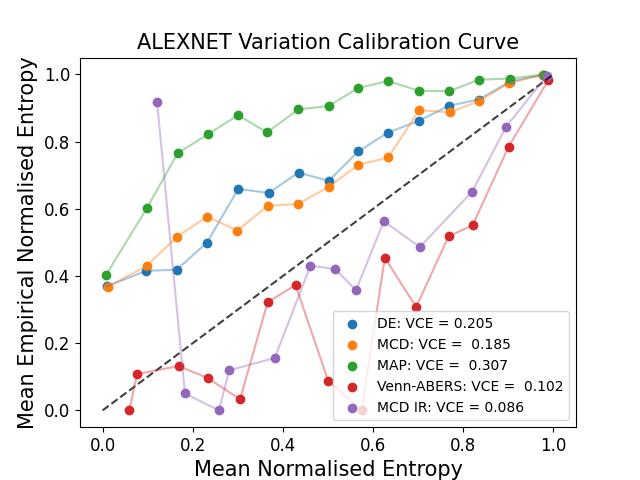}
    \includegraphics[width=0.4\linewidth]{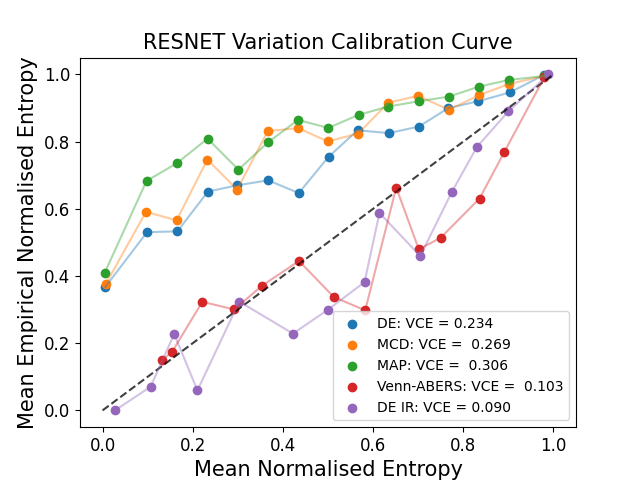}
    \includegraphics[width=0.4\linewidth]{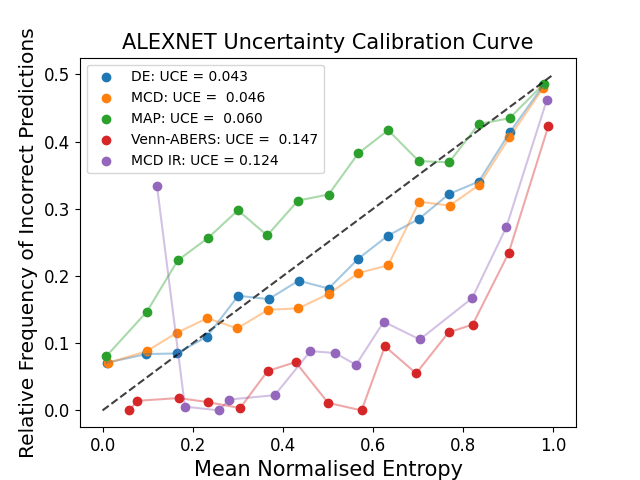}
    \includegraphics[width=0.4\linewidth]{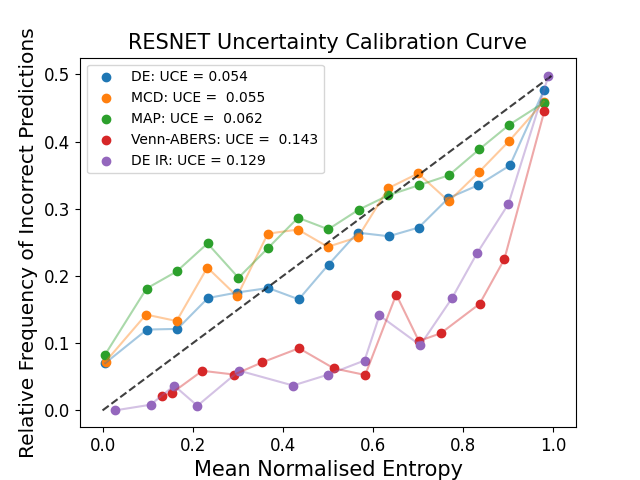}
    \caption{Reliability diagrams for ECE (top), VCE (middle), and UCE (bottom) calibration for 5 chosen UQ methods for both models. For \textbf{alexnet}, we include the results of Isotonic Regression (IR) on the MCD predictions, and for \textbf{resnet}, we include the results of IR on the DE  predictions. The black dashed line represents the ideal calibration relationship.}
    \label{fig:reliability_diagrams_all_UQ}
\end{figure}

The non-adaptive reliability diagrams in Figure~\ref{fig:reliability_diagrams_all_UQ} provide further insight into how the post-hoc recalibration influences uncertainty reliability. DE, MAP estimation, and MCD all follow similar trends, exhibiting overconfidence, where this is less severe for DE and MCD for \textbf{alexnet}. In comparison, the post-hoc methods are underconfident. 
For the MCD+IR and Venn-ABERS methods with \textbf{alexnet}, we find that smaller predicted entropies (i.e.\ more confident predictions) tend to represent correct predictions, where subtle changes in magnitude did not reflect corresponding changes in inaccuracy. These insights highlight the importance of analysing local reliability using reliability diagrams, since it can reveal magnitude-dependent variation in reliability including trends of over/under confidence in estimates. 

We suggest that if better \textbf{global} confidence calibration is desired, post-hoc calibration techniques may provide optimal performance. In addition, the Venn-ABERS method has statistical coverage guarantees, which are often desirable in constructing interpretable uncertainties. With that said, the \textbf{conditional/local} reliability results presented in Section \ref{sec:adaptive_per_class_eval} indicate a more nuanced picture of the potential benefits of using these post-hoc techniques. For all UQ methods, it is important to balance reliability with predictive performance; the NLL metric encapsulates this information, and can help with determining the most suitable UQ technique. 

Furthermore, the choice of hyperparameters for MCD and other techniques can have a significant effect on the quality of uncertainties. While here we compare the results of one implementation of each technique, in principle, several parametrisations should be used to find that which provides the best balance between predictive performance and uncertainty reliability. Indeed more stark differences in reliability between MCD and DE could be observed when using a larger dropout rate.

\begin{table}[!ht]
    \centering
    \footnotesize
    \begin{tabular}{|c|*{6}{|c}||c|}
    \hline
        \multirow{2}{*}{\textbf{UQ type}} & \multicolumn{7}{c|}{\textbf{Performance Metrics}} \\ \cline{2-8}
        & \textbf{ECE↓} & \textbf{ACE↓} & \textbf{smECE↓} & \textbf{UCE↓} & \textbf{VCE↓} & \textbf{NLL↓} & \textbf{AUC↑} \\ \hline\hline
        \textbf{MAP} & 0.122 &  0.122 & 0.113 & 0.060 & 0.307 & 1.153 & 0.81 \\ \hline
        \textbf{MCD} & 0.071 & 0.072 & 0.067 & 0.046 & 0.185 & 0.880 & 0.82 \\ \hline
        \textbf{DE} &  0.076 & 0.077 & 0.068 & \underline{\textbf{0.043}} & 0.205 & 0.927 & \textbf{0.83} \\ \hline
        \textbf{MCD+TS} & 0.068 & 0.071 & 0.063 & 0.143 & 0.103 & 0.749 & 0.82 \\ \hline
        \textbf{MCD+IR} & \underline{\textbf{0.048}} &  \underline{\textbf{0.057}} & \underline{\textbf{0.044}} & 0.124 & \underline{\textbf{0.086}} & 0.734 & 0.82 \\ \hline
        \textbf{DE+TS} & 0.081 & 0.085 & 0.072 & 0.148 & 0.118 & 0.734 & \textbf{0.83} \\ \hline
        \textbf{DE+IR} & 0.062 & 0.069 & 0.053 & 0.126 & 0.099 & \textbf{0.705} & \textbf{0.83} \\ \hline
        \textbf{Venn-ABERS} & 0.055 & 0.070 & 0.055 & 0.147 & 0.102 & 0.739 & 0.81 \\ \hline
    \end{tabular}
    \caption{\textbf{alexnet} evaluation metric results across all UQ methods. Best average calibration values for each metric are in bold. The best metric result across both models is underlined. Abbreviations are defined in Table \ref{tab:abb}.}
    \label{table:AF alexnet}
\end{table}

\begin{table}[!ht]
    \centering
    \footnotesize
    \begin{tabular}{|c|*{6}{|c}||c|}
    \hline
        \multirow{2}{*}{\textbf{UQ type}} & \multicolumn{7}{c|}{\textbf{Performance Metrics}} \\ \cline{2-8}
        & \textbf{ECE↓} & \textbf{ACE↓} & \textbf{smECE↓} & \textbf{UCE↓} & \textbf{VCE↓} & \textbf{NLL↓} & \textbf{AUC↑} \\ \hline\hline
        \textbf{MAP} & 0.098 & 0.100 & 0.086 & 0.062 & 0.306 & 1.169 & 0.84 \\ \hline
        \textbf{MCD} & 0.087 & 0.084 & 0.076 & 0.055 & 0.269 & 1.053 & 0.85 \\ \hline
        \textbf{DE} & 0.074 & 0.076 & 0.064 & \textbf{0.054} & 0.234 & 0.973 & \underline{\textbf{0.86}} \\ \hline
        \textbf{MCD+TS} & 0.078 & 0.078 & 0.073 & 0.158 & 0.110 & 0.691 & 0.85  \\ \hline
        \textbf{MCD+IR} & 0.055 & 0.075 & \textbf{0.044} & 0.133 & 0.095 & 0.712 & 0.85 \\ \hline
        \textbf{DE+TS} & 0.075 & 0.084 & 0.070 & 0.149 & 0.102 & 0.692 & \underline{\textbf{0.86}} \\ \hline
        \textbf{DE+IR} & \textbf{0.050} & \textbf{0.059} & \textbf{0.044} & 0.129 & \textbf{0.090} & \underline{\textbf{0.682}} & 0.85 \\ \hline
        \textbf{Venn-ABERS} & 0.055 & 0.063	& 0.048 & 0.143 & 0.103 & 0.691 & 0.84 \\ \hline
        
    \end{tabular}
    \caption{\textbf{resnet} model calibration metric results across all UQ methods. Best average calibration values for each metric are in bold. The best metric result across both models is underlined.}
    \label{table:AF resnet}
\end{table}

\subsubsection{Adaptive (per-class) evaluation}
\label{sec:adaptive_per_class_eval}
\begin{figure}
    \centering
    \includegraphics[width=0.245\linewidth]{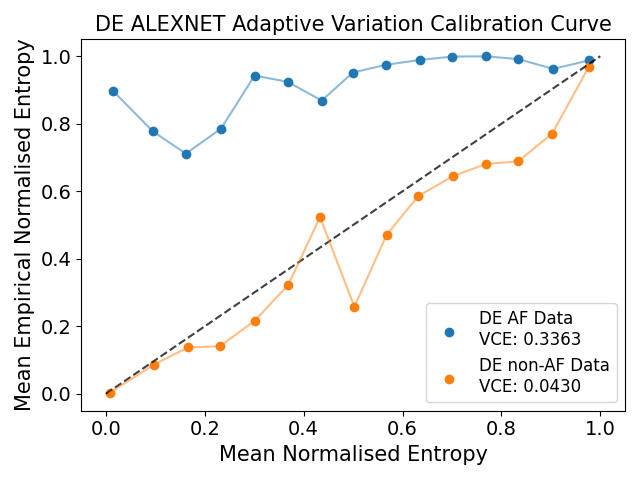}
    \includegraphics[width=0.245\linewidth]{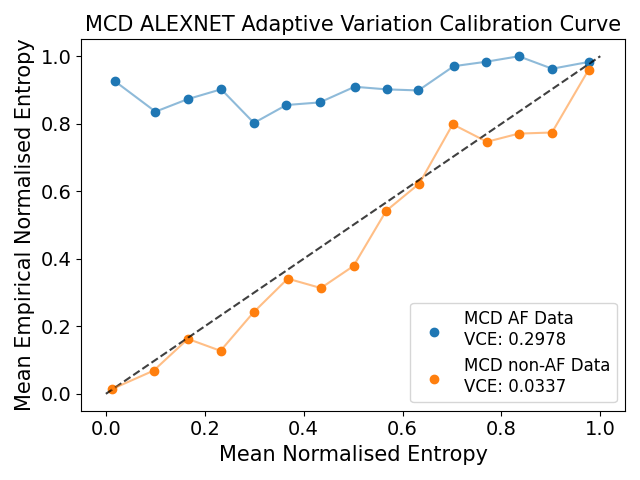}
    \includegraphics[width=0.245\linewidth]{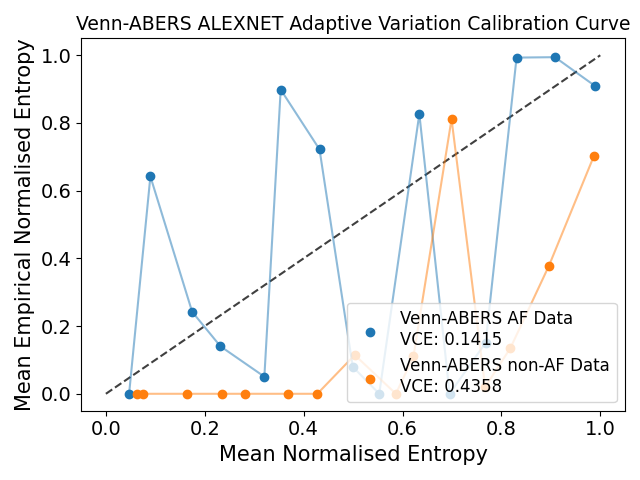}
    \includegraphics[width=0.245\linewidth]{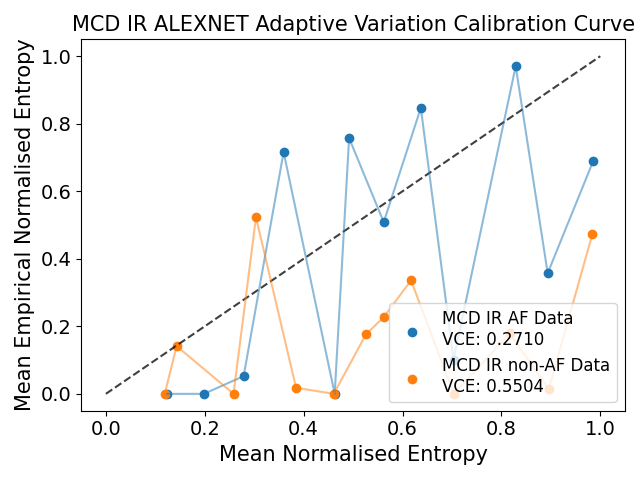}
    \includegraphics[width=0.245\linewidth]{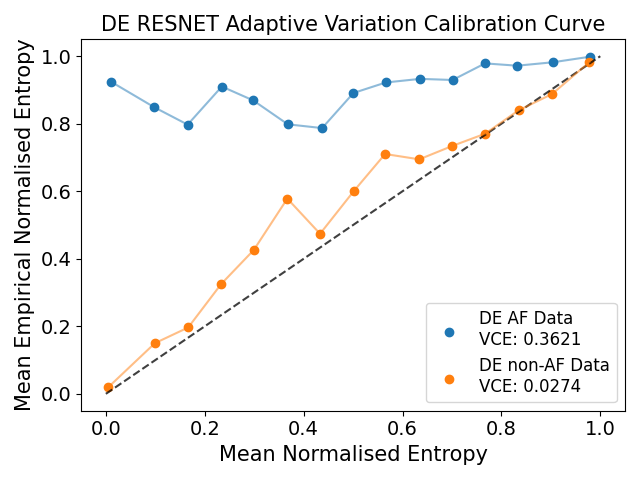}
    \includegraphics[width=0.245\linewidth]{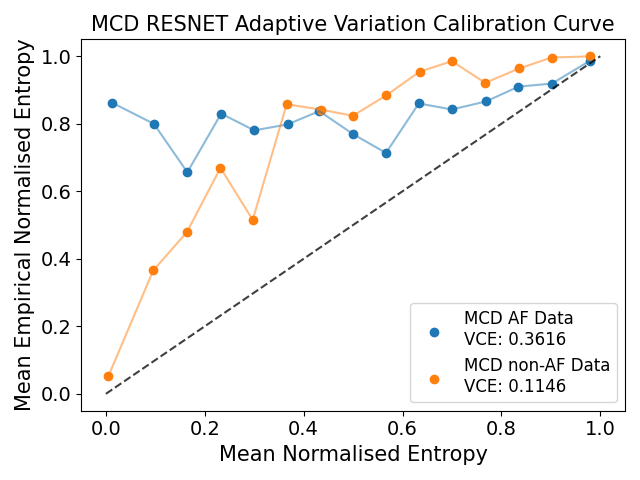}
    \includegraphics[width=0.245\linewidth]{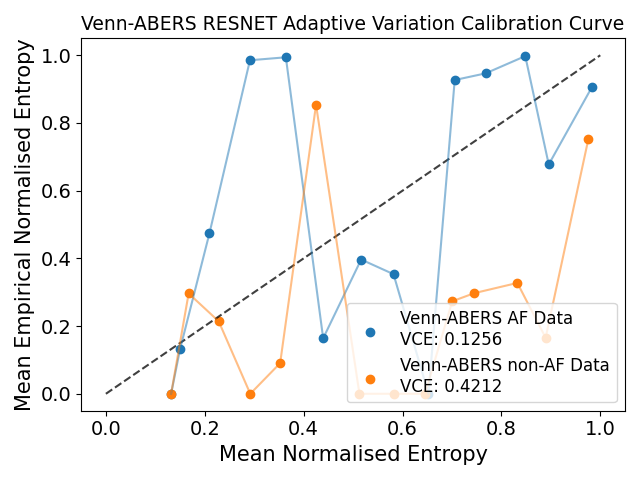}
    \includegraphics[width=0.245\linewidth]{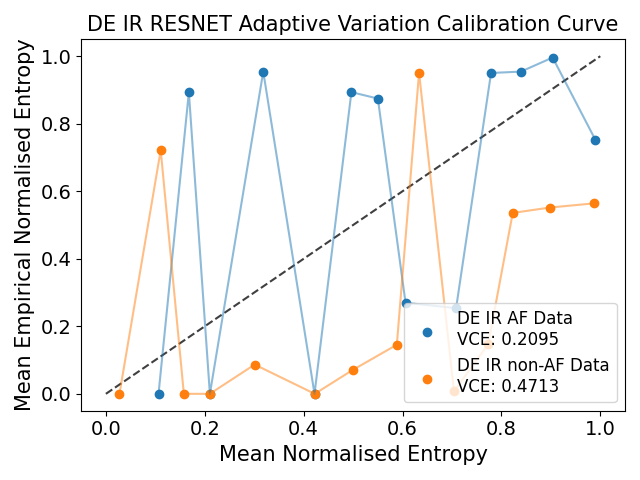}
    \caption{Adaptive variation calibration plots as assessed by the VCE for DE, MCD, Venn-ABERS, and IR for both models. Venn-ABERS results are given for both models, whilst MCD+IR results are shown for \textbf{alexnet}, and DE+IR results are shown for \textbf{resnet}.}
    \label{fig:adaptive_VCE_all_UQ}
\end{figure}

\begin{figure}[ht]
    \centering
    \includegraphics[width=0.245\linewidth]{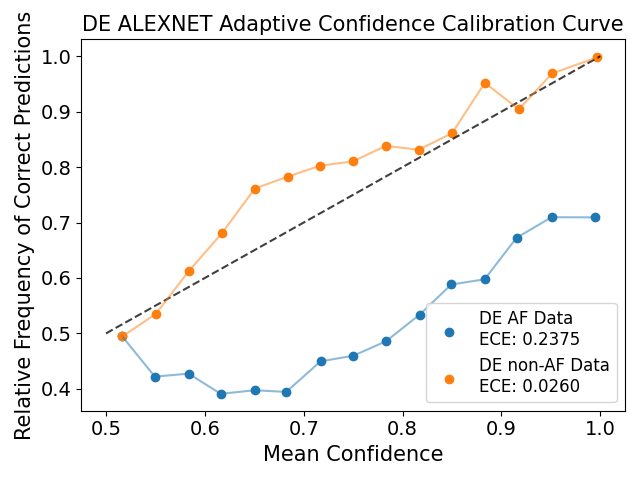}
    \includegraphics[width=0.245\linewidth]{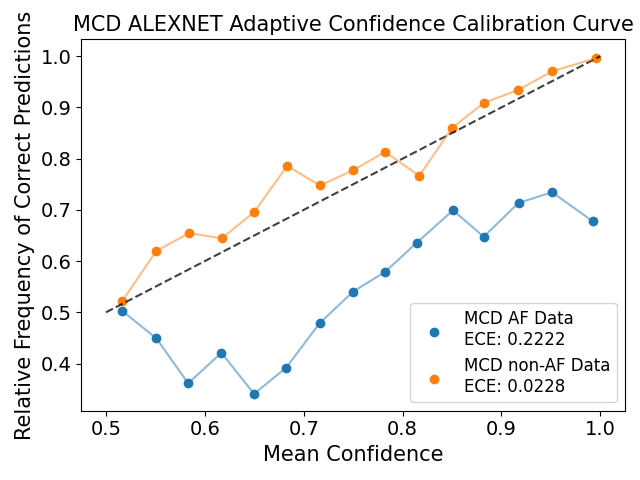}
    \includegraphics[width=0.245\linewidth]{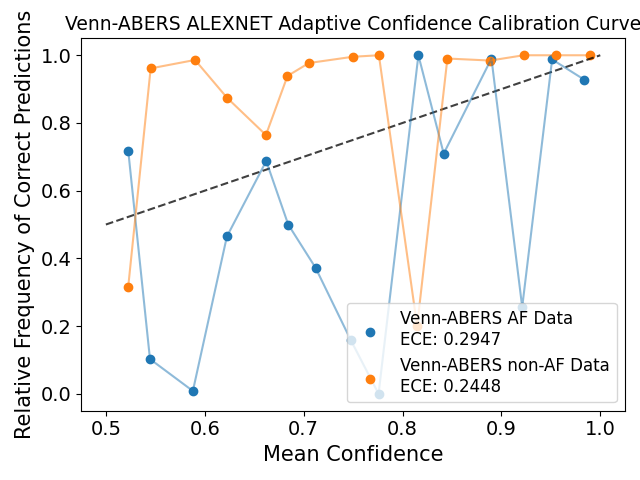}
    \includegraphics[width=0.245\linewidth]{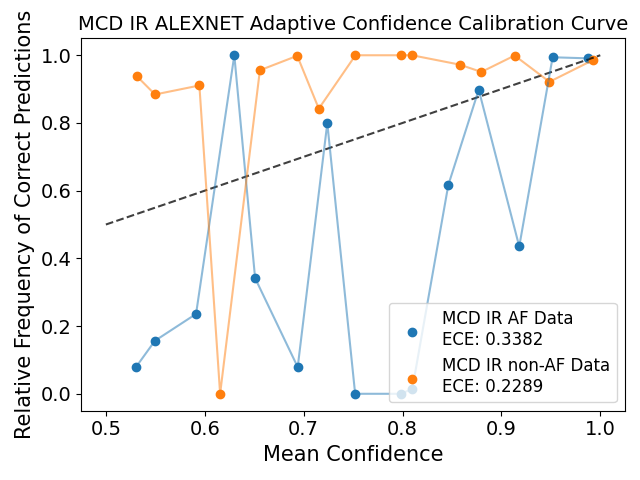}
    \includegraphics[width=0.245\linewidth]{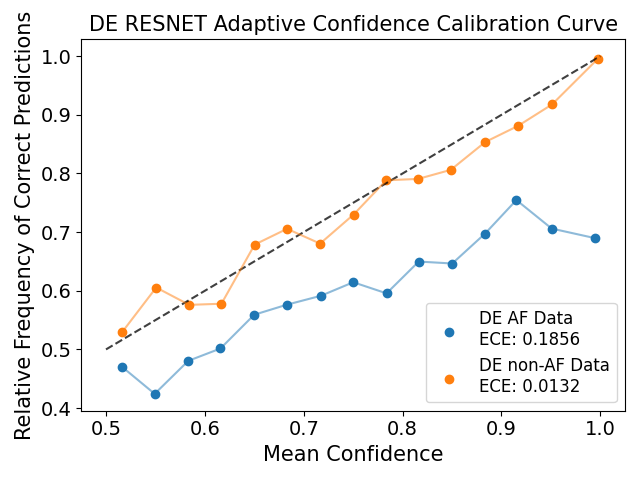}
    \includegraphics[width=0.245\linewidth]{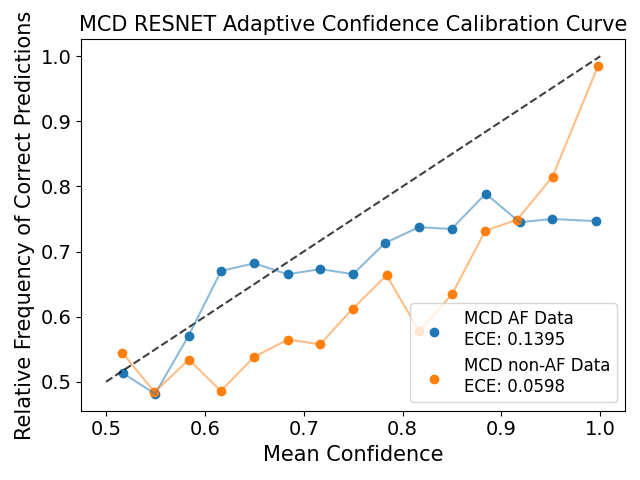}
    \includegraphics[width=0.245\linewidth]{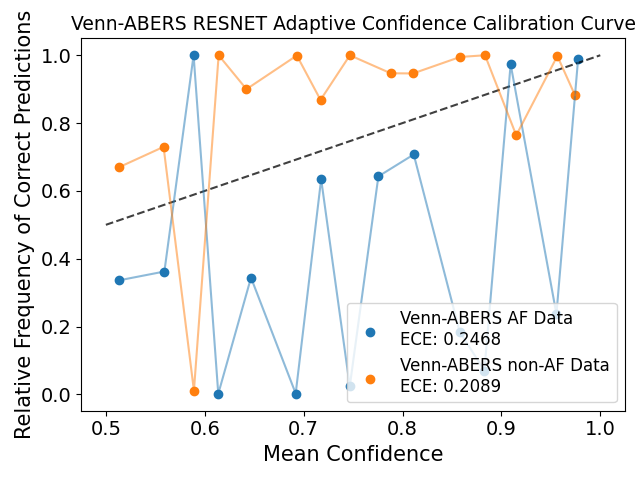}
    \includegraphics[width=0.245\linewidth]{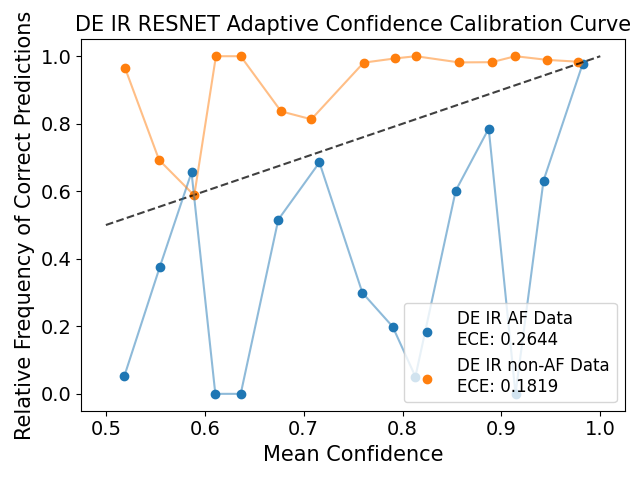}
    \caption{Adaptive reliability plots DE, MCD, Venn-ABERS, and IR for both models. Venn-ABERS results are given for both models, whilst MCD+IR results are shown for \textbf{alexnet}, and DE+IR results are shown for \textbf{resnet}.}
    \label{fig:adaptive_ECE_all_UQ}
\end{figure}

Here, we investigate differences between the UQ techniques when the metrics are assessed in the adaptive reliability setting, where in our case we choose to stratify the results based on the ground truth class, to assess uncertainty reliability on non-AF/AF results. Tables~\ref{table:per class AF alexnet} and \ref{table:per class AF resnet} show the per-class metric results for the \textbf{alexnet} and \textbf{resnet} models respectively, with results given as non-AF/AF. We also show per-class reliability diagrams for the ECE and VCE in Figures~\ref{fig:adaptive_VCE_all_UQ} and \ref{fig:adaptive_ECE_all_UQ} respectively.

\begin{table}[!ht]
    \centering
    \small
    \begin{tabular}{|c|*{6}{|c}||c|}
    \hline
        \textbf{UQ type} & \multicolumn{7}{c|}{\textbf{Performance Metrics}} \\ \cline{2-8}
        & \textbf{ECE↓} & \textbf{ACE↓} & \textbf{smECE↓} & \textbf{UCE↓} & \textbf{VCE↓} & \textbf{NLL↓} & \textbf{BA↑} \\ \hline\hline
        \textbf{MAP} & 0.102/0.158 & 0.235/\textbf{0.361} & 0.101/0.128 & \textbf{0.044}/0.117 & 0.159/0.390 & 0.645/1.992 & 0.77/\textbf{0.67} \\ \hline
        \textbf{MCD} & \textbf{0.023}/0.222 & 0.204/0.439 & \textbf{0.022}/0.203 & 0.089/0.134 & \textbf{0.034}/0.298 & 0.434/1.615 & 0.86/0.54 \\ \hline
        \textbf{DE} & 0.026/0.237 & \textbf{0.180}/0.436 & 0.024/0.210 & 0.087/0.153 & 0.043/0.336 & \textbf{0.378}/1.834 & 0.88/0.54 \\ \hline
        \textbf{MCD+TS} & 0.150/0.103 & 0.305/0.466 & 0.150/\textbf{0.086} & 0.236/\underline{\textbf{0.063}} & 0.303/0.074 & 0.576/1.035 & 0.86/0.54 \\ \hline
        \textbf{MCD+IR} & 0.229/0.338 & 0.283/0.470 & 0.223/0.294 & 0.343/0.278 & 0.550/0.271 & 0.524/1.081 & 0.95/0.36 \\ \hline
        \textbf{DE+TS} & 0.162/\textbf{0.099} & 0.292/0.464 & 0.162/\textbf{0.086} & 0.244/0.079 & 0.324/\underline{\textbf{0.064}} & 0.550/1.039 & 0.88/0.54 \\ \hline
        \textbf{DE+IR} & 0.238/0.346 & 0.277/0.458 & 0.226/0.299 & 0.349/0.269 & 0.572/0.305 & 0.501/1.044 & \underline{\textbf{0.96}}/0.37 \\ \hline
        \textbf{Venn-ABERS} & 0.245/0.295 & 0.310/0.462 & 0.206/0.197 & 0.309/0.171 & 0.436/0.142 & 0.574/\textbf{1.010} & 0.91/0.45 \\ \hline
    \end{tabular}
    \normalsize
    \caption{\textbf{alexnet} evaluation metric results across all UQ methods, conditioned on the ground truth class label. Results are given as non-AF/AF. The best metric value per class is in bold. The best metric result per class across both models is underlined.}
    \label{table:per class AF alexnet}
\end{table}

\begin{table}[!ht]
    \centering
    \small
    \begin{tabular}{|c|*{6}{|c}||c|}
    \hline
        \textbf{UQ type} & \multicolumn{7}{c|}{\textbf{Performance Metrics}} \\ \cline{2-8}
        & \textbf{ECE↓} & \textbf{ACE↓} & \textbf{smECE↓} & \textbf{UCE↓} & \textbf{VCE↓} & \textbf{NLL↓} & \textbf{BA↑} \\ \hline\hline
        \textbf{MAP} & 0.042/0.193 & 0.168/0.381 & 0.040/0.156 & 0.024/0.135 & 0.120/0.448 & 0.489/2.291 & 0.86/0.64 \\ \hline
        \textbf{MCD} & 0.060/0.139 & 0.182/\underline{\textbf{0.349}} & 0.057/0.106 & \underline{\textbf{0.015}}/0.135 & 0.115/0.362 & 0.512/1.949 & 0.83/\underline{\textbf{0.69}} \\ \hline
        \textbf{DE} & \underline{\textbf{0.013}}/0.186 & \underline{\textbf{0.158}}/0.403 & \underline{\textbf{0.012}}/0.157 & 0.047/0.112 & \underline{\textbf{0.027}}/0.362 & \underline{\textbf{0.366}}/1.977 & 0.88/0.61 \\ \hline
        \textbf{MCD+TS} & 0.090/0.116 & 0.276/0.431 & 0.090/0.100 & 0.178/0.160 & 0.248/0.110 & 0.530/0.959 & 0.83/\underline{\textbf{0.69}} \\ \hline
        \textbf{MCD+IR} & 0.207/0.288 & 0.255/0.441 & 0.178/0.212 & 0.295/0.202 & 0.486/0.137 & 0.472/1.110 & 0.94/0.48 \\ \hline
        \textbf{DE+TS} & 0.127/\underline{\textbf{0.083}} & 0.258/0.454 & 0.127/\underline{\textbf{0.061}} & 0.212/\textbf{0.095} & 0.285/\textbf{0.079} & 0.487/1.031 & 0.88/0.61 \\ \hline
        \textbf{DE+IR} & 0.182/0.264 & 0.253/0.431 & 0.181/0.243 & 0.306/0.198 & 0.471/0.210 & 0.482/1.012 & \textbf{0.95}/0.46 \\ \hline
        \textbf{Venn-ABERS} & 0.209/0.247 & 0.292/0.405 & 0.177/0.189 & 0.300/0.159 & 0.421/0.126 & 0.555/\underline{\textbf{0.915}} & 0.91/0.52 \\ \hline
    \end{tabular}
    \normalsize
    \caption{\textbf{resnet} evaluation metric results across all UQ methods, conditioned on the ground truth class label. Results are given as non-AF/AF. The best metric value per class is in bold. The best metric result per class across both models is underlined.}
    \label{table:per class AF resnet}
\end{table}

The best-performing method for non-AF predictions in terms of nearly all calibration metrics (with the exception of UCE) was \textbf{resnet} MCD, whilst the best-performing method for AF predictions in terms of all calibration metrics except for ACE and UCE was either \textbf{resnet} DE+TS or \textbf{alexnet} DE+TS. For example, the best-performing ECE scores were achieved by \textbf{resnet} DE for non-AF (0.013) and \textbf{resnet} DE+TS for AF (0.083). The radically different impression of relative performance given by UCE can be attributed to deficiencies in the metric, as discussed in Section~\ref{sec:global}. The ACE metric is optimised by the MAP estimation method for AF predictions for \textbf{alexnet}, which is a surprising result given that it does not model various sources of uncertainty.

While the global calibration metrics suggest the models produced well-calibrated uncertainties, the adaptive metrics and reliability diagrams show that reliability for each class is comparatively worse across all UQ methods. This is the case for both the estimated confidences and entropies. Across both models, in general we see that non-AF uncertainties are more reliable than AF uncertainties. The post-hoc methods, including the Venn-ABERS method, exhibit poor reliability for both non-AF and AF cases (which is expected given their optimisation targets prioritise global calibration). It has been shown that the IR method minimises the Kullback-Leibler (KL) divergence, which in our binary classification problem where the target distribution is fixed, is equal to the log likelihood up to a constant \cite{berta2024classifier}, and we see this behaviour for the global reliability results as the IR method achieves the best NLL for both models. However, when viewed in the adaptive scenario, we see that this no longer holds, as the best NLL values for non-AF predictions and AF predictions correspond to the \textbf{alexnet} DE (0.378) and \textbf{resnet} Venn-ABERS (0.915) results respectively. These results suggest that UQ techniques prioritise global/non-local reliability, and are biased to the dominant class, and suggests a need for more refined UQ techniques that exhibit better adaptivity/small scale reliability. 

The results for the DE and MCD methods are comparable for both classes, and therefore the hypothesised benefits of the DE method (i.e.\ that empirically, it may capture more of the variance in the true posterior distribution) \cite{fort2019deep, wilson2020bayesian} are not evidenced by our results. In terms of the best-performing model, the adaptive reliability diagrams qualitatively indicate that the \textbf{resnet} model for DE and MCD produced more reasonable local reliability estimates across both classes. This indicates that the best UQ method/model is dependent on the scale at which uncertainty reliability is assessed.

The post-hoc techniques focus on reducing overconfidence irrespective of the ground truth class label. While these methods show optimal global reliability, as demonstrated in Figures~\ref{fig:adaptive_VCE_all_UQ} and \ref{fig:adaptive_ECE_all_UQ}, they sometimes demonstrate poorer adaptive reliability. In the clinical use case, the preferable choice of UQ method may have poorer global reliability, but better adaptive reliability. 

\subsection{Blood pressure estimation}\label{sec:regression results}
Uncertainty evaluation results are shown in Tables \ref{table:alexnet_calib_results} - \ref{table:resnet_calibfree_results}. Section~\ref{sec:calib regression results} presents results for the \textbf{calib} dataset, and Section~\ref{sec:calibfree regression results} presents results for the \textbf{calibfree} dataset. The tables show results for both systolic and diastolic blood pressures, with results given as SBP/DBP. We also present results for predictive performance in the form of the mean absolute error (MAE) and the mean absolute scaled error (MASE). Conformal prediction methods applied to MAP estimation and QR are shown as Conformalised MAP (CMAP) and Conformalised Quantile Regression (CQR) respectively. The CMAP results shown are for the same model of the ensemble as that of the MAP estimation predictions.

\subsubsection{Calib dataset results}\label{sec:calib regression results}

The \textbf{calib} dataset results are shown for all metrics in Tables~\ref{table:alexnet_calib_results} and \ref{table:resnet_calib_results}, alongside variance reliability diagrams and small scale reliability assessments in Figures~\ref{fig:ENCE_calib_all_UQ} and \ref{fig:bivariate_hist_calib_DE_MCD}. 

\begin{table}[ht]
    \centering
    \begin{tabular}{|c|*{8}{|c}||c|}
    \hline
        \textbf{UQ type} & \multicolumn{7}{c|}{\textbf{Performance Metrics}} \\ \cline{2-8}
        & \textbf{CRPS↓} & \textbf{PICP 1$\sigma$} & \textbf{PICP 2$\sigma$} & \textbf{CCE↓} & \textbf{ENCE↓} & \textbf{MAE (mmHg)↓} & \textbf{MASE} \\ \hline\hline

        \textbf{MAP} &  7.48 / 4.86 & 0.959 / 0.968 & 0.979 / 0.980 & 0.011 / \textbf{0.001} & 0.102 / 0.103 & 10.53 / 6.83 &  0.71 / 0.72  \\ \hline
        
        \textbf{MCD} &  7.29 / \textbf{4.73} & 0.973 / 0.986 & 0.983 / 0.984 & \textbf{0.001} / 0.002 & 0.085 / 0.080 & 10.25 / \textbf{6.64}  & \textbf{0.69} / \textbf{0.70}\\ \hline
        
        \textbf{DE} &  \textbf{7.27} / \textbf{4.73} & 1.025 / 1.030 & 1.002 / 1.002 & 0.003 / \textbf{0.001} & 0.025 / 0.032 & 10.25 / 6.66  & \textbf{0.69} / \textbf{0.70} \\ \hline
        
        \textbf{QR 1$\sigma$} &  7.42 / 4.83 & 0.954 / 0.957 & n / a  & 0.004 / 0.002 & 0.116 / 0.138 & 10.42 / 6.77  & 0.70 / 0.71 \\ \hline
        \textbf{QR 2$\sigma$} &  7.53 / 4.87 & n / a & 0.980 / 0.979 & 0.023 / 0.011 & 0.105 / 0.117 & 10.63 / 6.87 & 0.71 / 0.72 \\ \hline
        
        \textbf{DE+TS} &  \textbf{7.27} / \textbf{4.73} & 1.021 / 1.027 & \textbf{1.000} / \textbf{1.001} & 0.003 / \textbf{0.001} & 0.026 / 0.033 & 10.25 / 6.66  & \textbf{0.69} / \textbf{0.70}\\ \hline
        \textbf{DE+IR} &  \textbf{7.27} / 4.74 & 1.028 / 1.030 & 1.002 / \textbf{1.001} & \textbf{0.001} / 0.002 & 0.028 / 0.034 & \textbf{10.23} / 6.66  & \textbf{0.69} / \textbf{0.70}\\ \hline
        
        \textbf{QR 1$\sigma$+TS} & 7.41 / 4.83 & 1.025 / 1.028 & \textbf{1.000} / 0.998 & 0.003 / \textbf{0.001} & 0.025 / 0.047 & 10.42 / 6.77 & 0.70 / 0.71 \\ \hline
        \textbf{QR 1$\sigma$+IR} & 7.41 / 4.82 & 1.032 / 1.031 & 1.001 / \textbf{0.999} & 0.002 / \textbf{0.001} & 0.022 / 0.043 & 10.41 / 6.77 & 0.70 / 0.71 \\ \hline
        
        \textbf{QR 2$\sigma$+TS} &  7.52 / 4.87 & 1.009 / 1.018 & 1.002 / 1.002 & 0.020 / 0.009 & 0.017 / 0.030 & 10.63 / 6.87 & 0.71 / 0.72 \\ \hline
        \textbf{QR 2$\sigma$+IR} &  7.44 / 4.85 & 1.022 / 1.022 & 1.002 / \textbf{1.001} & \textbf{0.001} / \textbf{0.001} & \textbf{0.016}  / \textbf{0.029} & 10.49 / 6.83 & 0.70 / 0.72 \\ \hline
        
        \textbf{CMAP 1$\sigma$} &  7.48 / 4.86 & \textbf{1.002} / 1.020 & n / a & 0.009 / \textbf{0.001} & 0.037 / 0.040 & \multirow{2}{*}{10.53 / 6.83}  & \multirow{2}{*}{0.71 / 0.72}\\ \cline{1-6}
        \textbf{CMAP 2$\sigma$} &  7.48 / 4.86 & n / a & 0.997 / 1.004 & 0.009 / 0.002 & 0.029 / 0.038 &  &  \\ \hline
        
        \textbf{CQR 1$\sigma$} &  7.41 / 4.83 & 0.994 / \textbf{1.012} & n / a & 0.002 / \textbf{0.001} & 0.052 / 0.051 & 10.42 / 6.77 & 0.70 / 0.71 \\ \hline
        \textbf{CQR 2$\sigma$} &  7.52 / 4.87 & n / a & 0.998 / 1.004 & 0.020 / 0.009 & 0.032 / 0.030 & 10.63 / 6.87  & 0.71 / 0.72\\ \hline
        
    \end{tabular}
    \caption{Regression uncertainty reliability and predictive performance metric results for the \textbf{alexnet} model on the \textbf{calib} dataset for systolic/diastolic blood pressure prediction. The MAE from predicting the median of the \textbf{calib} training set (a quantity used in the calculation of the MASE) is 14.91 mmHg for SBP and 9.52 mmHg for DBP. Abbreviations are defined in Table \ref{tab:abb}.}
    \label{table:alexnet_calib_results}
\end{table}

\begin{table}[ht]
    \centering
    \begin{tabular}{|c|*{8}{|c}||c|}
    \hline
        \textbf{UQ type} & \multicolumn{7}{c|}{\textbf{Performance Metrics}} \\ \cline{2-8}
        & \textbf{CRPS↓} & \textbf{PICP 1$\sigma$} & \textbf{PICP 2$\sigma$} & \textbf{CCE↓} & \textbf{ENCE↓} & \textbf{MAE (mmHg)↓} & \textbf{MASE} \\ \hline\hline
        \textbf{MAP} & 6.63 / 4.20 & 0.960 / 0.976 & 0.972 / 0.974 & 0.016 / 0.011 & 0.133 / 0.137 & 9.28 / 5.87  & 0.62 / 0.62 \\ \hline
        
        \textbf{MCD} & \textbf{5.99} / \textbf{3.81} & 1.085 / 1.094 & 1.008 / 1.009 & 0.008 / 0.012 & 0.062 / 0.076 & \textbf{8.38} / \textbf{5.32}  & \textbf{0.56} / \textbf{0.56} \\ \hline
        
        \textbf{DE} & 6.18 / 3.94 & 1.107 / 1.119 & 1.015 / 1.017 & 0.013 / 0.012 & 0.103 / 0.106 & 8.64 / 5.49  & 0.58 / 0.58\\ \hline
        
        \textbf{QR 1$\sigma$} & 6.73 / 4.32 & 0.955 / 0.951 & n / a & 0.005 / 0.002 & 0.139 / 0.167 & 9.38 / 6.02  & 0.63 / 0.63 \\ \hline
        \textbf{QR 2$\sigma$} & 6.78 / 4.36 & n / a & 0.981 / 0.977 & 0.020 / 0.004 & 0.096 / 0.110 & 9.49 / 6.10  & 0.64 / 0.64\\ \hline
        
        \textbf{DE+TS} & 6.16 / 3.92 & 1.031 / 1.048 & 0.996 / 0.997 & 0.007 / 0.004 & 0.082 / 0.093 & 8.64 / 5.49  & 0.58 / 0.58\\ \hline
        \textbf{DE+IR} & 6.16 / 3.92 & 1.047 / 1.056 & 0.998 / 0.998 & \textbf{0.002} / 0.003 & 0.084 / 0.099 & 8.62 / 5.48  & 0.58 / 0.58 \\ \hline

        \textbf{QR 1$\sigma$+TS} & 6.71 / 4.31 & 1.041 / 1.040 & 0.997 / 0.995 & 0.005 / 0.002 & \textbf{0.027} / \textbf{0.028} & 9.38 / 6.02 & 0.63 / 0.63 \\ \hline 
        \textbf{QR 1$\sigma$+IR} & 6.70 / 4.32 & 1.046 / 1.038 & 0.998 / 0.994 & \textbf{0.002} / 0.006 & 0.032 / 0.035 & 9.35 / 6.03 & 0.63 / 0.63  \\ \hline 
        
        \textbf{QR 2$\sigma$+TS} & 6.77 / 4.35 & 1.031 / 1.026 & 0.998 / 0.996 & 0.019 / 0.004 & 0.052 / 0.052 & 9.49 / 6.10  & 0.64 / 0.64\\ \hline
        \textbf{QR 2$\sigma$+IR} & 6.72 / 4.35 & 1.041 / 1.022 & 0.998 / 0.993 & \textbf{0.002} / 0.005 & 0.058 / 0.058 & 9.38 / 6.10  & 0.63 / 0.64\\ \hline
        
        \textbf{CMAP 1$\sigma$} & 6.62 / 4.20 & \textbf{1.001} / 1.004 & n / a & 0.014 / 0.010 & 0.073 / 0.096 & \multirow{2}{*}{9.28 / 5.87} & \multirow{2}{*}{0.62 / 0.62} \\ \cline{1-6}
        \textbf{CMAP 2$\sigma$} & 6.62 / 4.20 & n / a & \textbf{1.001} / \textbf{0.999} & 0.015 / 0.012 & 0.049 / 0.062 &   & \\ \hline
        
        \textbf{CQR 1$\sigma$} & 6.72 / 4.31 & 0.982 / \textbf{1.001} & n / a & 0.004 / \textbf{0.001} & 0.093 / 0.084 & 9.38 / 6.02  & 0.63 / 0.63\\ \hline
        \textbf{CQR 2$\sigma$} & 6.78 / 4.35 & n / a & \textbf{0.999} / 0.996 & 0.019 / 0.004 & 0.066 / 0.062 & 9.49 / 6.10 & 0.64 / 0.64 \\ \hline
        
    \end{tabular}
    \caption{Regression uncertainty reliability and predictive performance metric results for the \textbf{resnet} model on the \textbf{calib} dataset for systolic/diastolic blood pressure prediction. The MAE from predicting the median of the \textbf{calib} training set (a quantity used in the calculation of the MASE) is 14.91 mmHg for SBP and 9.52 mmHg for DBP.}
    \label{table:resnet_calib_results}
\end{table}

\afterpage{
\begin{landscape}
    \begin{figure}[b]
        \centering
        \includegraphics[width=0.24\linewidth, height=0.15\linewidth, keepaspectratio]{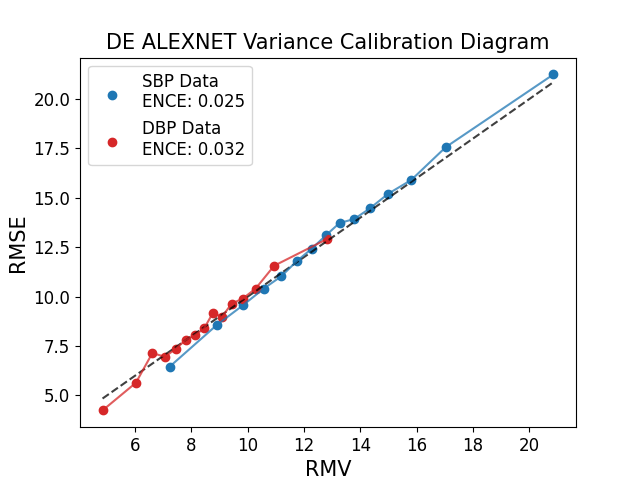}
        \includegraphics[width=0.24\linewidth, height=0.15\linewidth, keepaspectratio]{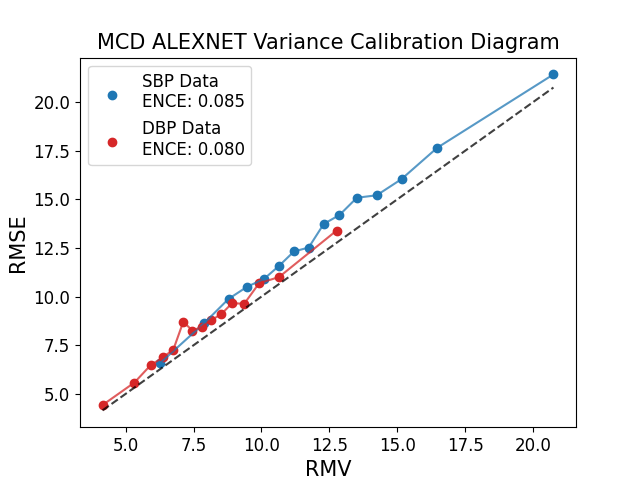}
        \includegraphics[width=0.24\linewidth, height=0.15\linewidth, keepaspectratio]{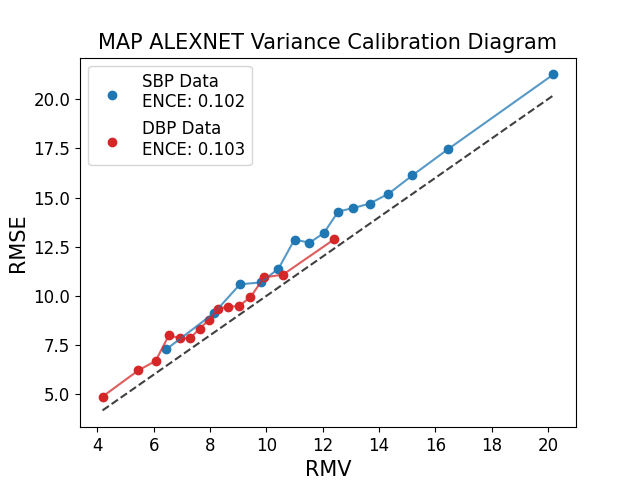}
        \includegraphics[width=0.24\linewidth, height=0.15\linewidth, keepaspectratio]{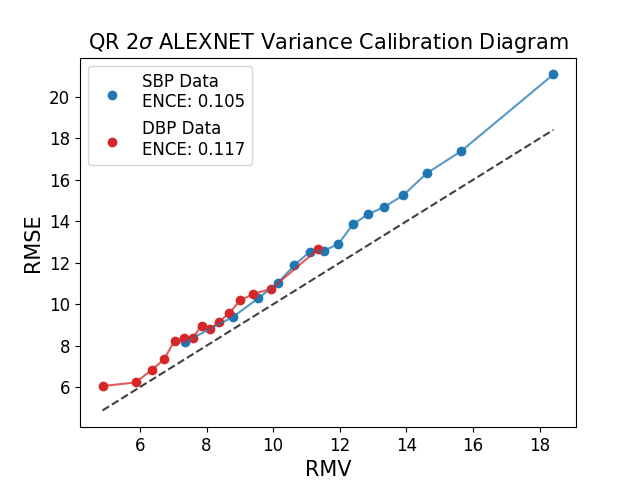} \\
        \includegraphics[width=0.24\linewidth, height=0.15\linewidth, keepaspectratio]{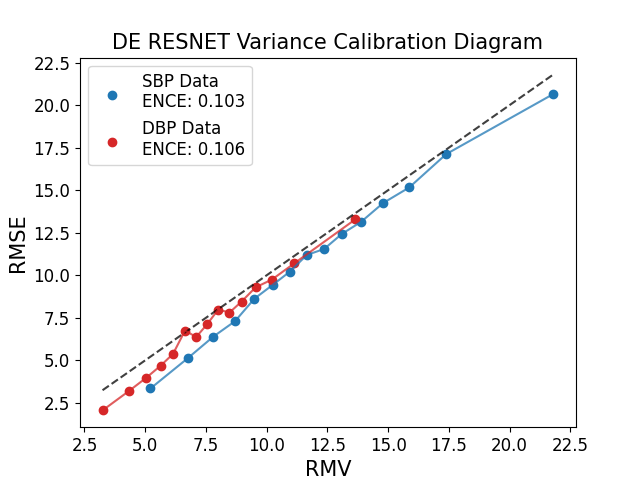}
        \includegraphics[width=0.24\linewidth, height=0.15\linewidth, keepaspectratio]{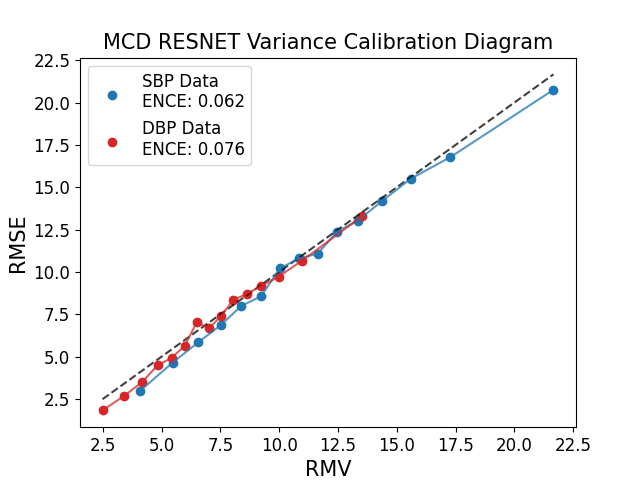}
        \includegraphics[width=0.24\linewidth, height=0.15\linewidth, keepaspectratio]{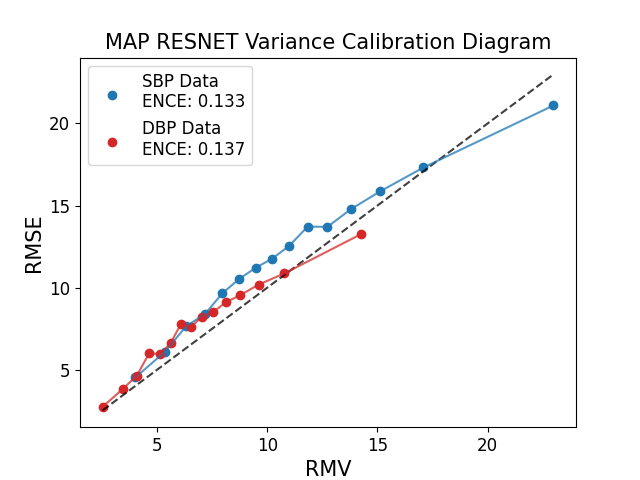}
        \includegraphics[width=0.24\linewidth, height=0.15\linewidth, keepaspectratio]{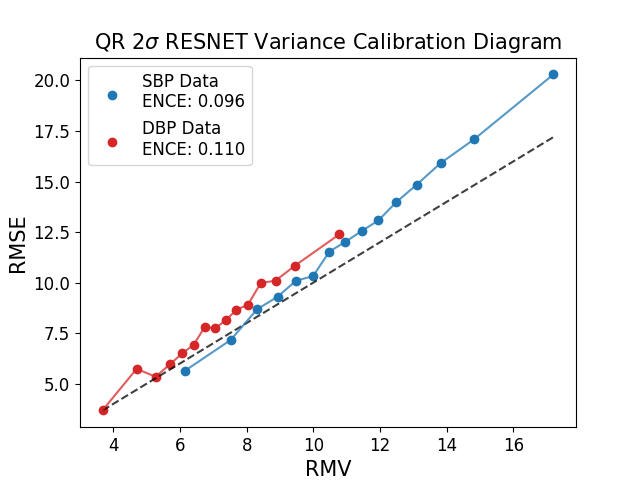}
        \caption{ENCE reliability diagrams of the 4 main UQ methods (DE, MCD, MAP, and QR) prior to recalibration for \textbf{alexnet} (top) and \textbf{resnet} (bottom) for the \textbf{calib} dataset. The quantile regression results are shown for the $2\sigma$ confidence level.}
        \label{fig:ENCE_calib_all_UQ}
    \end{figure}

    \begin{figure}[t]
        \centering
        \includegraphics[width=0.24\linewidth, height=0.15\linewidth, keepaspectratio]{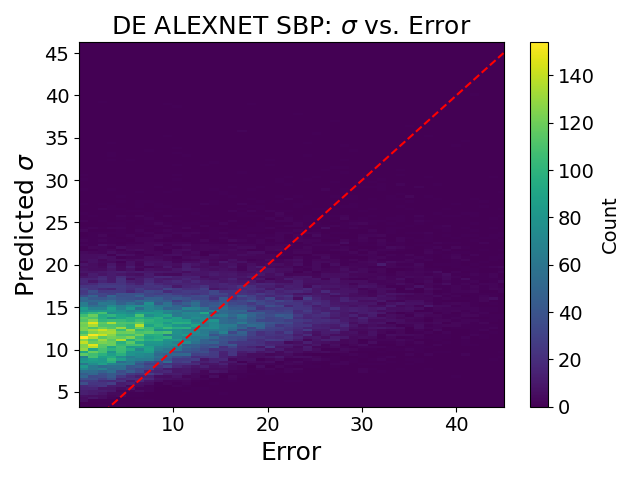}
        \includegraphics[width=0.24\linewidth, height=0.15\linewidth, keepaspectratio]{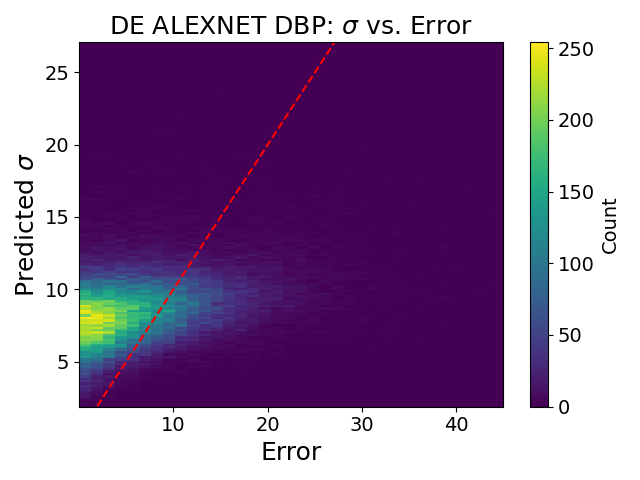}
        \includegraphics[width=0.24\linewidth, height=0.15\linewidth, keepaspectratio]{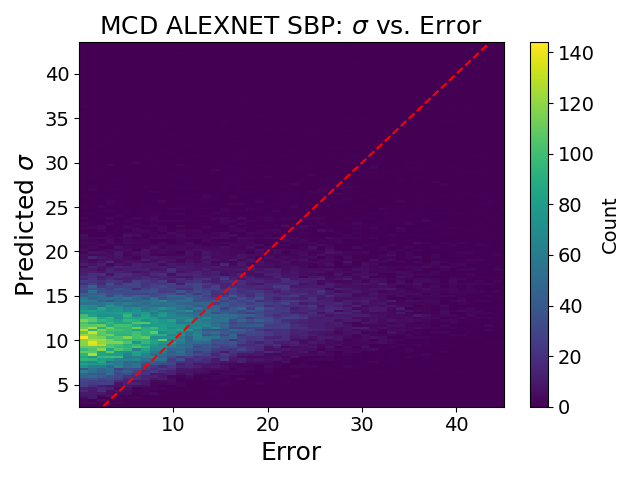}
        \includegraphics[width=0.24\linewidth, height=0.15\linewidth, keepaspectratio]{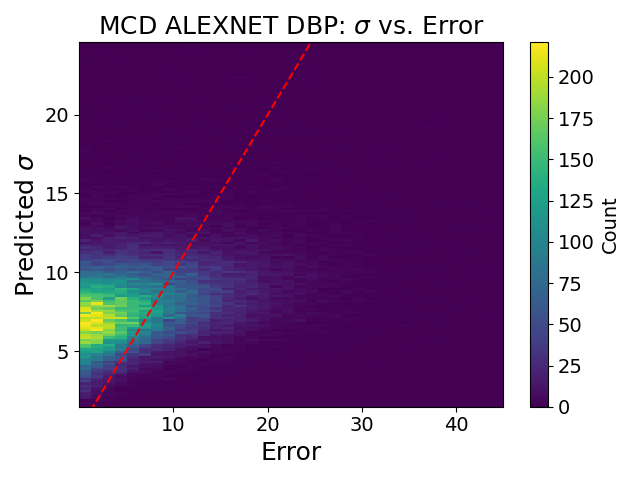} \\
        \includegraphics[width=0.24\linewidth, height=0.15\linewidth, keepaspectratio]{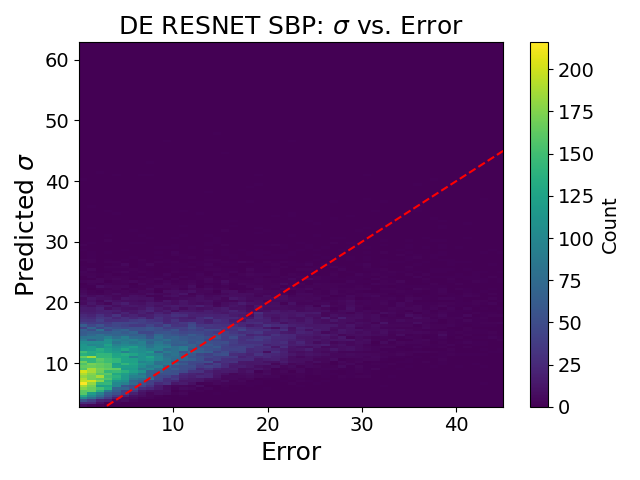}
        \includegraphics[width=0.24\linewidth, height=0.15\linewidth, keepaspectratio]{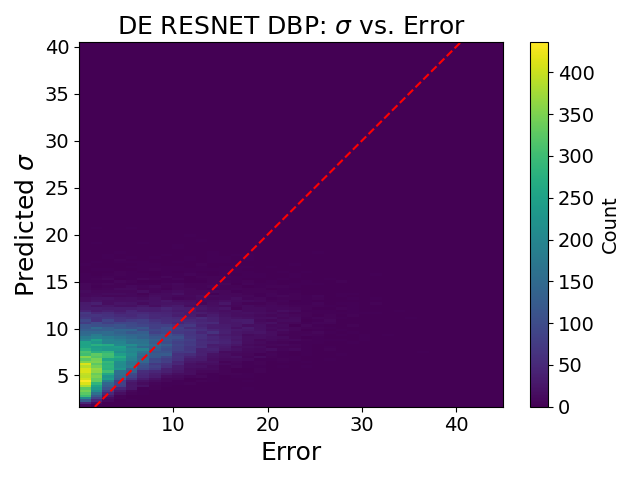}
        \includegraphics[width=0.24\linewidth, height=0.15\linewidth, keepaspectratio]{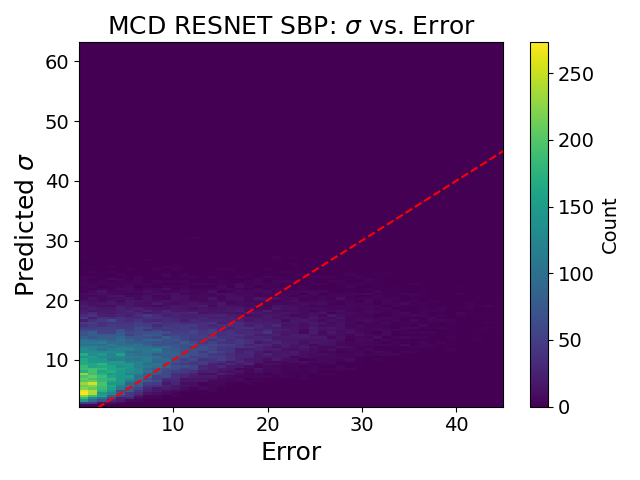}
        \includegraphics[width=0.24\linewidth, height=0.15\linewidth, keepaspectratio]{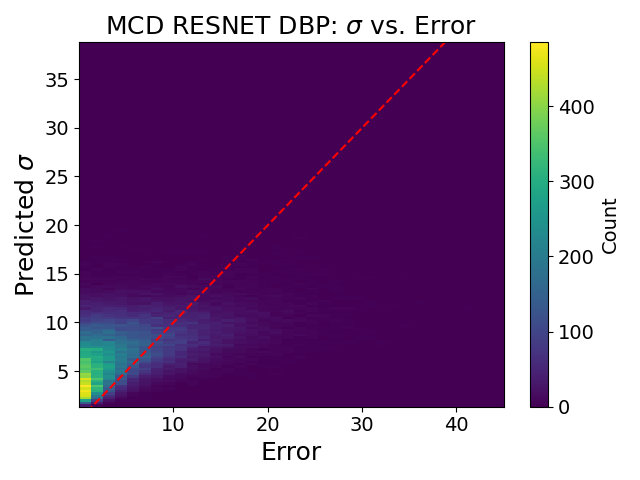}
        \caption{Bivariate histogram plots for \textbf{alexnet} and \textbf{resnet} models on the \textbf{calib} dataset for both SBP and DBP predictions for DE and MCD.}
        \label{fig:bivariate_hist_calib_DE_MCD}
    \end{figure}
\end{landscape}
}

We find notable differences in predictive performance between the SBP and DBP regression tasks as indicated by the MAE and CRPS. SBP prediction exhibits poorer predictive performance for both the \textbf{calib} and \textbf{calibfree} cases for both models. We hypothesise that the smaller variance of true DBP values may contribute to improved performance. However, we note that the scaled MASE is very similar in all cases for SBP and DBP prediction. Different trends are observed when considering other metrics for uncertainty reliability where SBP tends to exhibit better reliability across both models on the \textbf{calib} dataset according to the ENCE, but not for the others. These results demonstrates how assessing small scale reliability (ENCE) compared to global reliability (CRPS) can help draw more nuanced insights about model behaviour. Indeed, we find that the ENCE reliability diagrams show that the distribution of uncertainties and residuals for SBP are wider than those of DBP.

For the \textbf{calib} task, the \textbf{resnet} model achieved the highest predictive accuracy for both SBP and DBP. The MCD method (\textbf{resnet}) attained the highest predictive performance out of all the UQ techniques, with an MAE of 8.38 mmHg, and 5.32 mmHg for SBP and DBP respectively. In addition, results for the CRPS metric show the same trend as for the MAE, with the \textbf{resnet} model outperforming the \textbf{alexnet} model for all UQ methods, and the MCD method performing the best with a CRPS of 5.99 and 3.81 for SBP and DBP respectively. While the greater capacity of the \textbf{resnet} model improves predictive performance, its effect on uncertainty reliability is more mixed. This suggests that capacity does not provide clear advantages for all methods/metrics.

Considering local reliability by observing the ENCE values in Tables~\ref{table:alexnet_calib_results} and \ref{table:resnet_calib_results} and the reliability diagrams in Figure~\ref{fig:ENCE_calib_all_UQ} and bivariate histograms in Figure \ref{fig:bivariate_hist_calib_DE_MCD}; no single UQ method completely outperforms the others for both models/tasks. Figure~\ref{fig:ENCE_calib_all_UQ} indicates how the DE method improves reliability relative to the MAP estimation method, which we attribute to its capacity to capture model variance. From the same figure, we also find that the QR results for the $2\sigma$ level may suggest the method used to convert from intervals to distributions is not completely effective, as we see underestimated uncertainties at higher prediction errors. We hypothesise this may emerge due to the Gaussian assumption not holding for large prediction errors. The bivariate histograms (Figure \ref{fig:bivariate_hist_calib_DE_MCD}) show that there is large variance in the prediction errors, indicating that these methods do not achieve small-scale reliability. For \textbf{calib}, we find that larger model capacity does increase the number of well-calibrated smaller-magnitude uncertainties, but scale does not seem to consistently improve all reliability metrics.

For DE and MCD, uncertainty reliability varied considerably depending on the model and the chosen metric. Yet, both UQ methods had similar predictive performances. This may be attributed to the inherent similarities of their optimisation/evaluation; both methods were trained with the same loss, where the dropout rate for MCD was set to a low value of 5 \%, and both implement variants of ensemble averaging during inference.

An expected behaviour that we observe from the \textbf{calib} dataset results is that conformal prediction techniques improve interval coverage (PICP) after recalibration, when compared to the pre-calibrated quantile regression results. The PICP results for conformalised quantile regression (CQR) for both models are closer to the ideal value (i.e.\ the ratio of the obtained interval coverage to the target coverage value is closer to 1) after computing the rescaled quantiles using the calibration dataset. The IR and TS recalibration methods exhibit improved interval coverage at the $2\sigma$ confidence level, however they overestimate uncertainties at the $1\sigma$ level, given the fractional PICP values of greater than $1$. These methods also improve local reliability, with an improvement in ENCE for both models after quantile regression results are recalibrated. Due to the underlying Gaussian assumption when converting intervals to distribution parameters, there was no change in the MAE between pre- and post-calibrated quantile regression results, as the means (in this case, midpoints of the interval) remain unchanged (we do not use the median prediction, i.e.\ 0.5 quantile, to evaluate MAE as we want to compare the estimated means between UQ methods that output intervals).

Overall, the \textbf{calib} dataset BP results highlight the importance of assessing reliability at all three levels (individual, local/conditional, and average), as the relative performance of the UQ methods depend on the scale at which one assesses reliability. 

\subsubsection{Calibfree dataset results}\label{sec:calibfree regression results}

The \textbf{calibfree} dataset results are shown in Tables~\ref{table:alexnet_calibfree_results} and \ref{table:resnet_calibfree_results}, alongside the reliability diagrams in Figures~\ref{fig:ENCE_calibfree_all_UQ} and \ref{fig:bivariate_hist_calibfree_DE_MCD}.

\begin{table}[ht]
    \centering
    \begin{tabular}{|c|*{8}{|c}||c|}
    \hline
        \textbf{UQ type} & \multicolumn{7}{c|}{\textbf{Performance Metrics}} \\ \cline{2-8}
        & \textbf{CRPS↓} & \textbf{PICP 1$\sigma$} & \textbf{PICP 2$\sigma$} & \textbf{CCE↓} & \textbf{ENCE↓} & \textbf{MAE (mmHg)↓} & \textbf{MASE} \\ \hline\hline
        \textbf{MAP} & 8.75 / 5.60 & 0.997 / 1.010 & 1.010 / 1.006 & \textbf{0.0011} / 0.0068 & 0.046 / 0.052 & 12.44 / 7.94 & 0.84 / \textbf{0.84} \\ \hline
        \textbf{MCD} & 8.78 / 5.64 & 0.981 / \textbf{1.005} & \textbf{0.998} / 1.005 & 0.0054 / 0.0019 & 0.053 / 0.046 & 12.43 / 7.98 & 0.84 / 0.85\\ \hline
        \textbf{DE} & \textbf{8.72} / \textbf{5.59} & \textbf{0.990} / 1.012 & 1.004 / 1.007 &\textbf{0.0011} / 0.0004 & \textbf{0.045} / 0.048 & \textbf{12.37} / \textbf{7.92}  & \textbf{0.83} / \textbf{0.84} \\ \hline
        
        \textbf{QR 1$\sigma$} & 8.82 / 5.69 & 0.970 / 0.988 & n / a & 0.0072 / 0.0039 & 0.059 / 0.053 & 12.46 / 8.05  & 0.84 / 0.85\\ \hline
        \textbf{QR 2$\sigma$} & 8.82 / 5.68 & n / a & 0.991 / \textbf{0.996} & 0.0039 / 0.0010 & 0.067 / \textbf{0.045} & 12.48 / 8.04 & 0.84 / 0.85\\ \hline
        
        \textbf{DE+TS} & 8.74 / 5.60 & 0.935 / 0.958 & 0.987 / 0.990 & 0.0040 / \textbf{0.0001} & 0.095 / 0.069 & \textbf{12.37} / \textbf{7.92} & \textbf{0.83} / \textbf{0.84}\\ \hline
        \textbf{DE+IR} & 8.75 / 5.63 & 0.930 / 0.950 & 0.984 / 0.985 & 0.0070 / 0.0140 & 0.104 / 0.079 & \textbf{12.37} / 7.97 & \textbf{0.83} / 0.85\\ \hline
        
        \textbf{QR 1$\sigma$+TS} & 8.83 / 5.69 & 0.937 / 0.956 & 0.986 / 0.986 & 0.0091 / 0.0052 & 0.094 / 0.080 & 12.46 / 8.05 & 0.84 / 0.85 \\ \hline 
        \textbf{QR 1$\sigma$+IR} & 8.83 / 5.69 & 0.938 / 0.956 & 0.986 / 0.986 & 0.0086 / 0.0049 & 0.093 / 0.080 & 12.46 / 8.05 &  0.84 / 0.85 \\ \hline
        
        \textbf{QR 2$\sigma$+TS} & 8.82 / 5.68 & 0.938 / 0.976 & 0.986 / 0.991 & 0.0048 / 0.0013 & 0.086 / 0.056 & 12.48 / 8.04 & 0.84 / 0.85\\ \hline
        \textbf{QR 2$\sigma$+IR} & 8.82 / 5.70 & 0.932 / 0.965 & 0.985 / 0.988 & 0.0072 / 0.0090 & 0.098 / 0.069 & 12.46 / 8.07 & 0.84 / 0.86\\ \hline
        
        \textbf{CMAP 1$\sigma$} & 8.78 / 5.61 & 0.917 / 0.929 & n / a & 0.0050 / 0.0100 & 0.115 / 0.102 & \multirow{2}{*}{12.44 / 7.94} & \multirow{2}{*}{0.84 / \textbf{0.84}}\\ \cline{1-6}
        \textbf{CMAP 2$\sigma$} & 8.76 / 5.62 & n / a & 0.995 / 0.977 & 0.0027 / 0.0111 & 0.067 / 0.117 &  & \\ \hline
        
        \textbf{CQR 1$\sigma$} & 8.86 / 5.70 & 0.890 / 0.929 & n / a & 0.0142 / 0.0070 & 0.168 / 0.113 & 12.46 / 8.05 & 0.84 / 0.85\\ \hline
        \textbf{CQR 2$\sigma$} & 8.81 / 5.69 & n / a & 0.996 / 0.972 & 0.0032 / 0.0043 & 0.047 / 0.115 & 12.48 / 8.04 & 0.84 / 0.85\\ \hline
    \end{tabular}
    \caption{Regression uncertainty reliability and predictive performance metric results for the \textbf{alexnet} model on the \textbf{calibfree} dataset for systolic/diastolic blood pressure prediction. The MAE from predicting the median of the \textbf{calibfree} training set (a quantity used in the calculation of the MASE) is 14.87 mmHg for SBP and 9.43 mmHg for DBP. Abbreviations are defined in Table \ref{tab:abb}.}
    \label{table:alexnet_calibfree_results}
\end{table}

\begin{table}[ht]
    \centering
    \begin{tabular}{|c|*{8}{|c}||c|}
    \hline
        \textbf{UQ type} & \multicolumn{7}{c|}{\textbf{Performance Metrics}} \\ \cline{2-8}
        & \textbf{CRPS↓} & \textbf{PICP 1$\sigma$} & \textbf{PICP 2$\sigma$} & \textbf{CCE↓} & \textbf{ENCE↓} & \textbf{MAE (mmHg)↓} & \textbf{MASE} \\ \hline\hline
        \textbf{MAP} & 8.81  /  5.68 & 0.951  /  0.971 & 0.986  /  0.990 & 0.018  /  0.023 & 0.109  /  0.077 & 12.41  /  8.04 & \textbf{0.83} / 0.85\\ \hline
        \textbf{MCD} & 8.85  /  5.59 & 0.987  /  1.024 & \textbf{1.001}  /  1.007 & 0.008  /  0.003 & 0.058  /  \textbf{0.027} & 12.54  /  7.89  & 0.84 / 0.84\\ \hline
        \textbf{DE} & \textbf{8.71}  /  5.57 & \textbf{1.000}  /  1.025 & 1.005  /  1.007 & \textbf{0.006}  /  0.007 & \textbf{0.041}  /  0.034 & \textbf{12.32}  /  7.86  & \textbf{0.83} / 0.83\\ \hline
        \textbf{QR 1$\sigma$} & 8.91  /  \textbf{5.54} & 0.951 / \textbf{0.999} & n / a & \textbf{0.006}  /  \textbf{0.001} & 0.099  /  0.046 & 12.59  /  \textbf{7.80}  & 0.85 / 0.83\\ \hline
        \textbf{QR 2$\sigma$} & 9.12  /  5.55 & n / a & 0.993 / \textbf{0.998} & 0.041  /  0.006 & 0.079  /  0.030 & 12.93  /  7.84  & 0.87 / 0.83\\ \hline
        \textbf{DE+TS} & 8.72  /  5.57 & 0.930  /  0.956 & 0.983  /  0.987 & 0.009  /  0.008 & 0.107  /  0.070 & \textbf{12.32}  /  7.86 & \textbf{0.83} / 0.83\\ \hline
        \textbf{DE+IR} & 8.72  /  5.57 & 0.931  /  0.956 & 0.983  /  0.987 & 0.010  /  0.008 & 0.106  /  0.070 & \textbf{12.32}  /  7.86 & \textbf{0.83} / 0.83\\ \hline
        
        \textbf{QR 1$\sigma$+TS} & 8.92 / \textbf{5.54} & 0.933 / 0.982 & 0.983 / 0.988 & 0.007 / \textbf{0.001} & 0.112 / 0.060 & 12.59 / \textbf{7.80} & 0.85 / \textbf{0.82} \\ \hline 
        \textbf{QR 1$\sigma$+IR} & 8.93 / 5.63 & 0.930 / 0.962 & 0.977 / 0.980 & 0.009 / 0.032 & 0.128 / 0.084 & 12.56 / 7.93 & 0.84 / 0.84 \\ \hline 
        
        \textbf{QR 2$\sigma$+TS} & 9.12  /  5.55 & 0.935  /  0.996 & 0.989  /  0.995 & 0.042  /  0.006 & 0.088  /  0.037 & 12.93  /  7.84 & 0.87 / 0.83\\ \hline
        \textbf{QR 2$\sigma$+IR} & 8.96  /  5.67 & 0.928  /  0.945 & 0.978  /  0.979 & 0.008  /  0.043 & 0.122  /  0.095 & 12.62  /  7.99 & 0.85 / 0.85 \\ \hline
        
        \textbf{CMAP 1$\sigma$} & 8.88  /  5.69 & 0.863 / 0.939 & n / a & 0.028  /  0.025 & 0.231  /  0.108 & \multirow{2}{*}{12.41  /  8.04} & \multirow{2}{*}{\textbf{0.83} / 0.85} \\ \cline{1-6}
        \textbf{CMAP 2$\sigma$} & 8.81  /  5.70 & n / a & 0.998 / 0.969 & 0.016  /  0.028 & 0.084  /  0.133 &  & \\ \hline
        \textbf{CQR 1$\sigma$} & 8.95  /  \textbf{5.54} & 0.891 / 0.958 & n / a & 0.012  /  0.002 & 0.173  /  0.089 & 12.59  /  \textbf{7.80} & 0.85 / 0.83\\ \hline
        \textbf{CQR 2$\sigma$} & 9.12  /  5.56 & n / a & 0.992 / 0.982 & 0.041  /  0.008 & 0.081  /  0.081 & 12.93  /  7.84 & 0.87 / 0.83\\ \hline
    \end{tabular}
    \caption{Regression uncertainty reliability and predictive performance metric results for the \textbf{resnet} model on the \textbf{calibfree} dataset for systolic/diastolic blood pressure prediction. The MAE from predicting the median of the \textbf{calibfree} training set (a quantity used in the calculation of the MASE) is 14.87 mmHg and 9.43 mmHg for DBP.}
    \label{table:resnet_calibfree_results}
\end{table}

In contrast to the \textbf{calib} task, no particular model provides consistently superior predictive performance (as shown by the CRPS and MAE). Related work suggests the \textbf{calibfree} task is challenging \cite{moulaeifard2025machine}. However like the \textbf{calib} case, the DBP prediction task exhibited superior predictive performance relative to SBP, although again with very similar MASE values. 

All intrinsic/post-hoc ensemble methods achieve similar results, with DE mostly marginally outperforming MAP estimation and MCD according to the ENCE and CRPS, suggesting that DE gives better uncertainty estimates with the chosen training hyperparameters. With that said, the reliability of uncertainty estimates is known to vary depending on the choice of hyperparameters; a grid search would be required to assess which UQ method provides optimal performance. Indeed, other works have highlighted that the dropout rate for MCD requires careful adjustment to acquire more calibrated uncertainty estimates, for example in \cite{verdoja2020notes}. 

Figures~\ref{fig:ENCE_calibfree_all_UQ} and \ref{fig:bivariate_hist_calibfree_DE_MCD} show the variance-based reliability/calibration diagrams and small-scale calibration bivariate histograms respectively for a selection of the results for the \textbf{calibfree} prediction task. The small-scale/local calibration of the selected UQ methods for \textbf{calibfree} predictions exhibit more variation in local trends of over and under confident predictions relative to the \textbf{calib} case (where the ENCE values improve for some UQ methods). Neither architecture exhibits better local calibration compared to the other. The qualitative assessment of small scale reliability in the bivariate histograms in  Figure~\ref{fig:bivariate_hist_calibfree_DE_MCD} for DE and MCD show poor reliability, and that the \textbf{calibfree} dataset BP prediction task is more challenging than that of the \textbf{calib} dataset (also reported in \cite{moulaeifard2025machine}), which may suggest that the difficulty of the prediction task can influence reliability of uncertainty estimates.

Post-hoc recalibration methods and conformal prediction for the \textbf{calibfree} results for quantile regression show a decrease in PICP. This is unexpected behaviour, especially for CQR, as the method is formulated such that on test (unseen) data, the intervals should on average obtain the defined coverage level of the ground truth values. One possible explanation for poor performance is that the calibration and test set are not sufficiently similar in the \textbf{calibfree} case, where there is no overlap in patients across each set.

These results point to potential systematic concerns in either the implementation of the recalibration methods or the assumed exchangeability across training, validation, calibration, and test datasets. Consequently, we cannot draw definitive conclusions regarding the efficacy of different UQ methods for the \textbf{calibfree} task. This highlights the importance of robustly verifying that the distributions are consistent across dataset splits, especially in the context of clinical use, as this can significantly impact model performance and the reliability of different UQ methods.

\begin{landscape}    
    \begin{figure}[b]
        \centering
        \includegraphics[width=0.24\linewidth, height=0.15\linewidth, keepaspectratio]{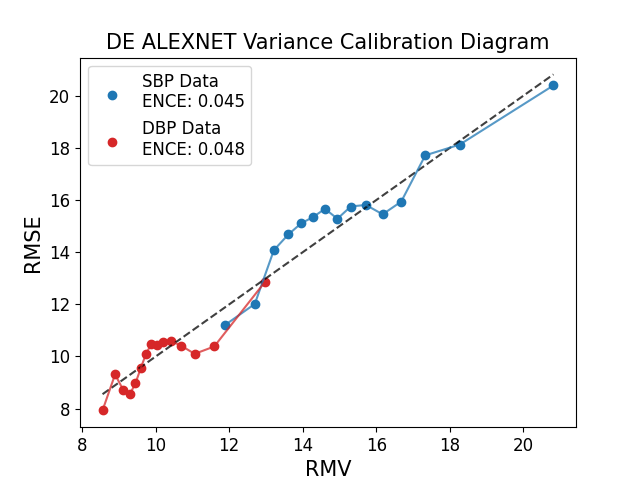}
        \includegraphics[width=0.24\linewidth, height=0.15\linewidth, keepaspectratio]{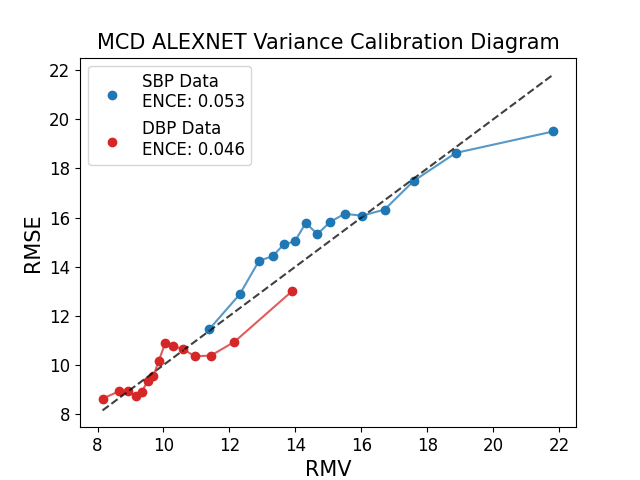}
        \includegraphics[width=0.24\linewidth, height=0.15\linewidth, keepaspectratio]{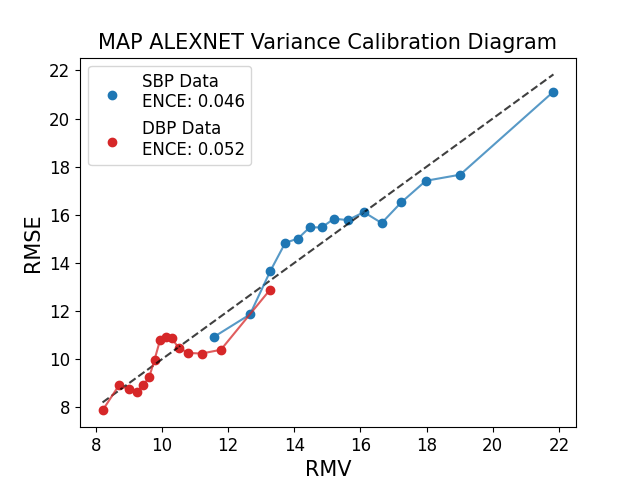}
        \includegraphics[width=0.24\linewidth, height=0.15\linewidth, keepaspectratio]{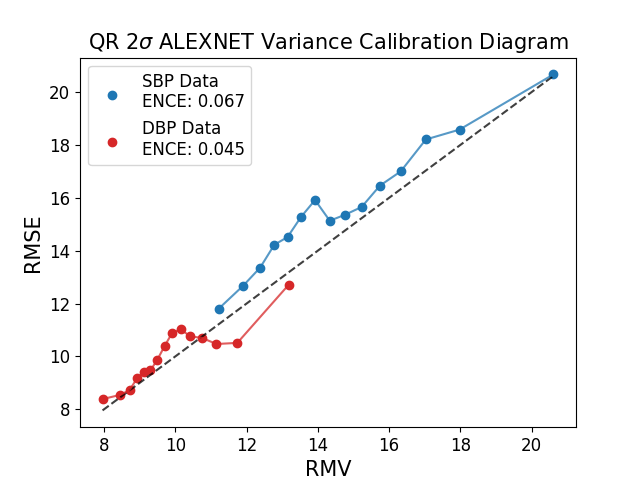} \\
        \includegraphics[width=0.24\linewidth, height=0.15\linewidth, keepaspectratio]{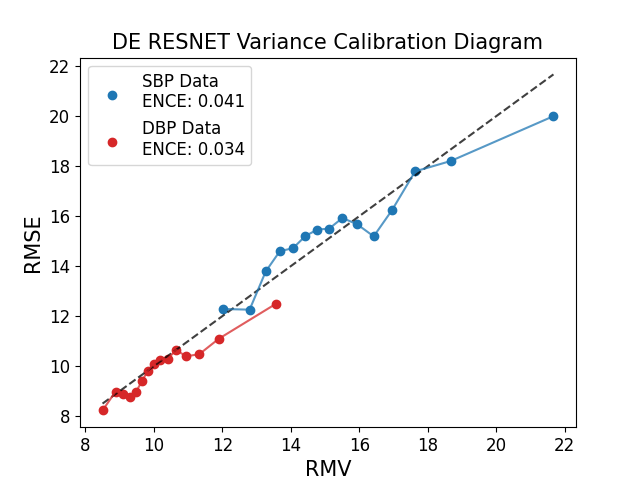}
        \includegraphics[width=0.24\linewidth, height=0.15\linewidth, keepaspectratio]{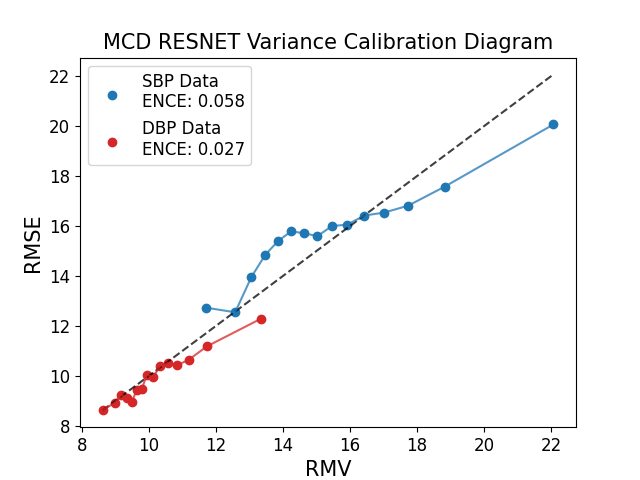}
        \includegraphics[width=0.24\linewidth, height=0.15\linewidth, keepaspectratio]{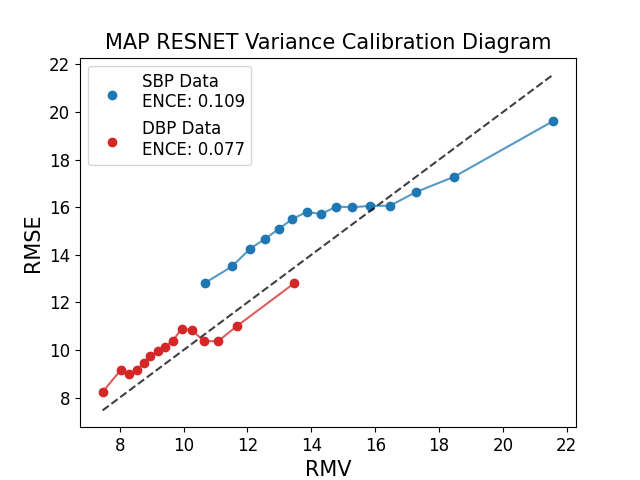}
        \includegraphics[width=0.24\linewidth, height=0.15\linewidth, keepaspectratio]{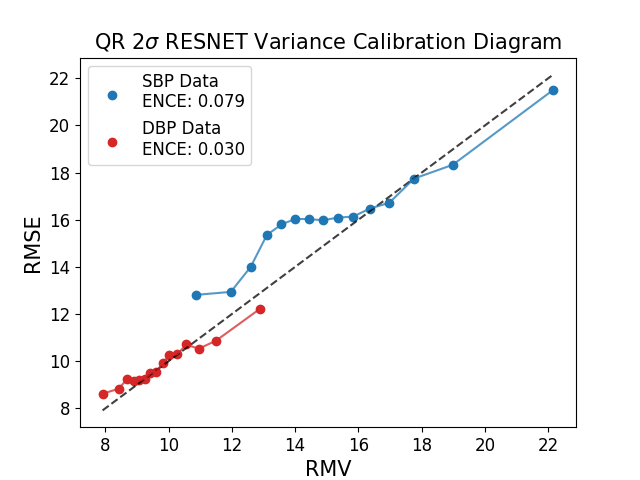}
        \caption{ENCE reliability diagrams of the 4 main UQ methods (DE, MCD, MAP, and QR) prior to recalibration for \textbf{alexnet} (top) and \textbf{resnet} (bottom) for the \textbf{calibfree} dataset. The quantile regression results are shown for the $2\sigma$ confidence level.}
        \label{fig:ENCE_calibfree_all_UQ}
    \end{figure}

    \begin{figure}[t]
        \centering
        \includegraphics[width=0.24\linewidth, height=0.15\linewidth, keepaspectratio]{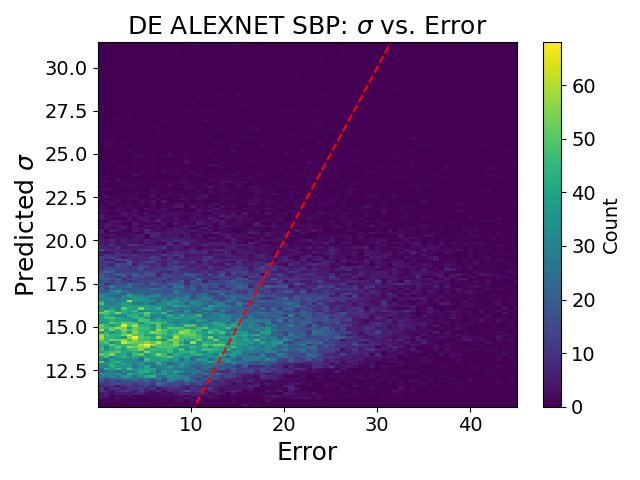}
        \includegraphics[width=0.24\linewidth, height=0.15\linewidth, keepaspectratio]{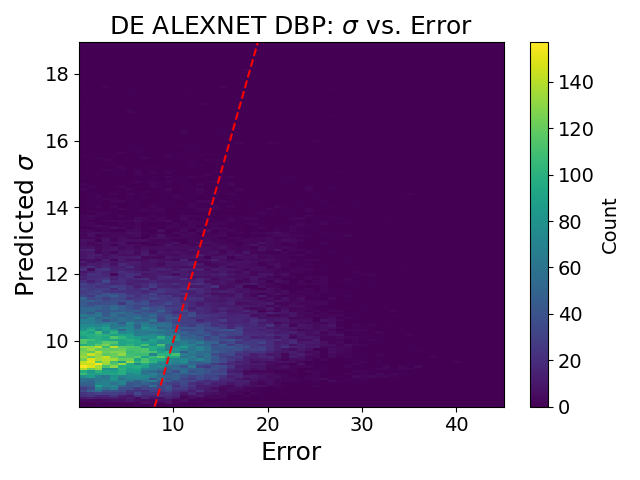}
        \includegraphics[width=0.24\linewidth, height=0.15\linewidth, keepaspectratio]{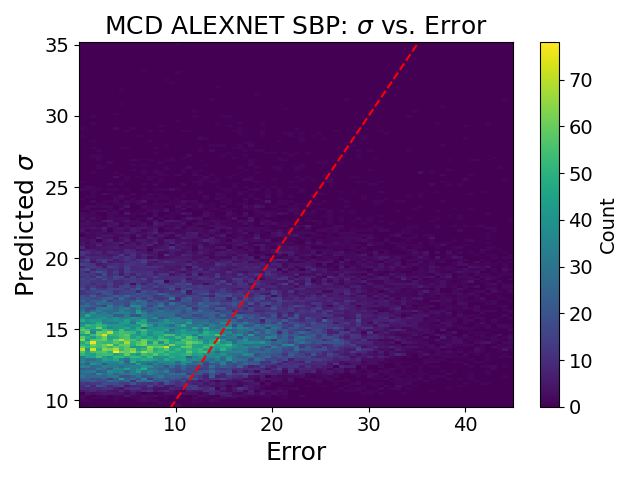}
        \includegraphics[width=0.24\linewidth, height=0.15\linewidth, keepaspectratio]{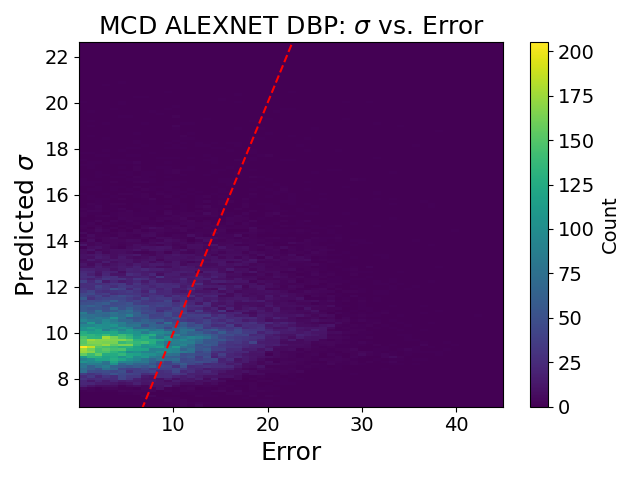} \\
        \includegraphics[width=0.24\linewidth, height=0.15\linewidth, keepaspectratio]{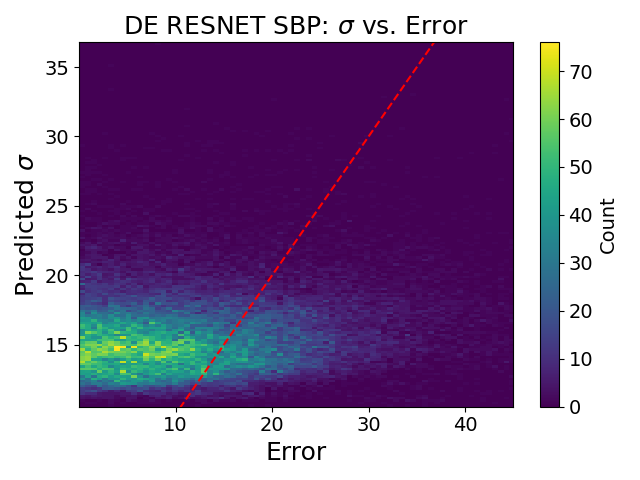}
        \includegraphics[width=0.24\linewidth, height=0.15\linewidth, keepaspectratio]{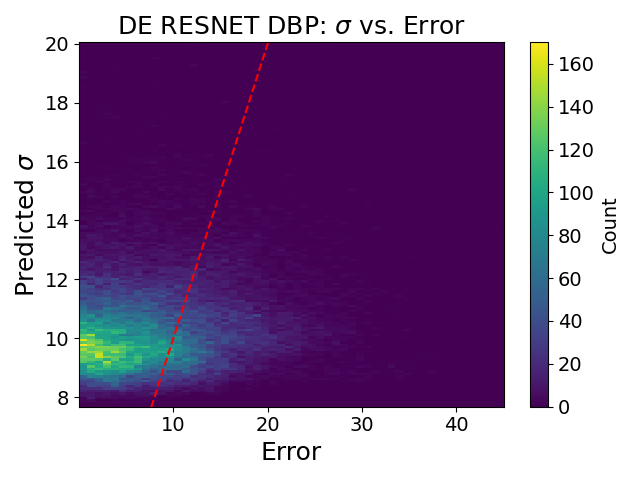}
        \includegraphics[width=0.24\linewidth, height=0.15\linewidth, keepaspectratio]{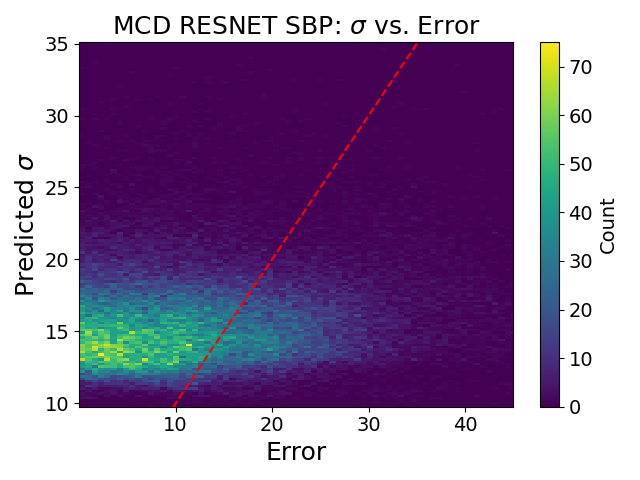}
        \includegraphics[width=0.24\linewidth, height=0.15\linewidth, keepaspectratio]{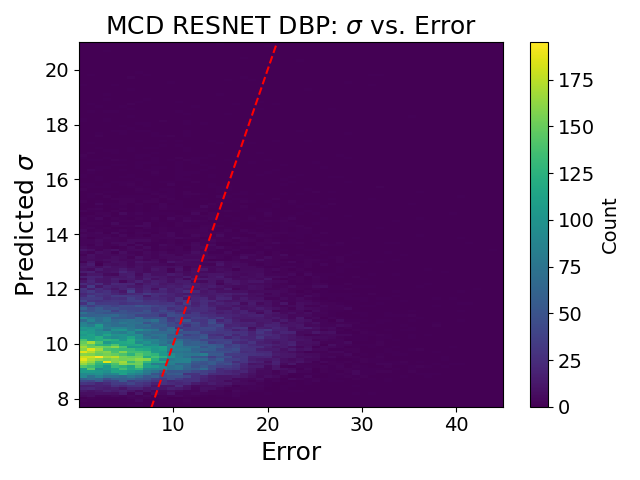}
        \caption{Bivariate histogram plots for \textbf{alexnet} and \textbf{resnet} models on the \textbf{calibfree} dataset for both SBP and DBP predictions for DE and MCD. Models exhibiting small scale reliability should have predictions present around the red line.}
        \label{fig:bivariate_hist_calibfree_DE_MCD}
    \end{figure}    
\end{landscape}

\subsection{General observations}
It is evident that given evaluation metrics cater to particular expressions of uncertainty, and that the reliability of predicted uncertainties is highly dependent upon the chosen expression of uncertainty and evaluation metric. Furthermore, any trade-offs between uncertainty reliability and predictive performance should be taken into account when deciding on the preferred UQ technique.

The results from the AF classification task indicate that the chosen UQ techniques do not result in adaptive uncertainty estimates (i.e.\ uncertainty reliability assessed per class), instead prioritising global reliability with a bias towards the dominant class. In particular, while the post-hoc recalibration techniques result in high global reliability for the class probabilities, the corresponding adaptive per-class metrics indicate poor reliability (which is not surprising as the predictions were not calibrated to achieve adaptive reliability). This emphasises the need to use evaluation metrics that cater to the practical use-case of the model. Indeed, individual reliability is the most effective measure of assessing uncertainty reliability, as this reflects how the model is likely to be used in practice, though there is a lack of robust quantitative metrics to assess this. It also points to a general need for UQ techniques that are sensitive to class imbalance, and which result in high adaptivity across classes. 

In general, uncertainty reliability varies considerably across the different techniques depending on the scale at which uncertainty reliability is assessed. Our qualitative assessment of uncertainty reliability at smaller scales through binning-based reliability diagrams revealed these biases in uncertainty reliability, emphasising the practical utility of this approach. These plots also revealed local trends of whether uncertainties were over or under estimated, which may aid in developing strategies for using estimates to inform diagnosis. Resnet-DE and ResNet-DE-TS produced the most optimally calibrated uncertainties (with statistical significance assessed via empirical bootstrapping)  according to our selection criteria based on the average of the per-class ECEs (a metric that considers adaptivity and local reliability).

We draw similar conclusions for the BP regression tasks, where the optimal UQ technique depended both on the chosen metric and the scale at which uncertainty reliability was assessed. Our own use of bivariate histograms to qualitatively assess small scale reliability revealed poor reliability for all models/UQ techniques, where the \textbf{calibfree} task exhibited the worst reliability. The ENCE's respective reliability diagrams help determine whether there are local trends of over or under confidence. The post-hoc recalibration techniques were observed to improve uncertainty reliability for the \textbf{calib} task for some models and some metrics (mostly for QR and \textbf{resnet}). For the \textbf{calibfree} task, post-hoc recalibration techniques did not generally improve uncertainty reliability according to the chosen metrics. We have also highlighted challenges with converting across different output types that should be resolved to enable more meaningful comparative studies across various UQ techniques. The nature of the \textbf{calibfree} dataset/task may have violated assumptions underpinning some of the UQ techniques; it is important to be wary of whether the task/data is compatible with a given UQ technique. Using the mean of the SBP and DBP ENCE (considering local reliability) as our selection criteria, we found that the top three models that produced the most reliable uncertainties for \textbf{calib} were alexnet-QR2$\sigma$-IR, alexnet-QR2$\sigma$-TS, and resnet-QR1$\sigma$-TS while the top three for \textbf{calibfree} were alexnet-DE, resnet-DE, and resnet-MCD (these methods did not exhibit statistically significant differences in uncertainty reliability).

\subsubsection{Limitations}
While not investigated here, uncertainty reliability will also depend on the parametrisation of each model/technique (e.g.\ the number of ensembles for DE, or the dropout rate for MCD). Ideally, a set of parametrisations should be grid searched when trying to choose an optimal UQ metric. Indeed, there were few consistent trends in the effect the choice of model had on uncertainty reliability.

\section{Conclusions and recommendations}
Generally, our recommendation for applying UQ to deep learning models is as follows: an optimal UQ technique may be chosen by first choosing the desired expression of uncertainty for a given task and then by evaluating adaptive and local versions of the relevant uncertainty reliability metrics for a range of model parametrisations. The technique/parametrisation that produces the best trade-off between adaptive and/or small-scale reliability and predictive performance should be implemented in practice. Reliability diagrams should be used to observe whether models exhibit local trends of over or under predicting estimates, and this can also be used as a factor for choosing an optimal UQ strategy if deemed important for the practical use case of the model. Modelling both aleatoric and epistemic uncertainty appears to provide better overall uncertainty reliability, and so these methods are generally preferable. Future work should consider ways to develop: confidence thresholds for keeping/rejecting an estimate, UQ techniques that encourage adaptivity, and quantitative individual reliability metrics.
\section*{Code availability}
A software repo is in preparation.
\section*{Data availability}
Scripts for data preprocessing are available at \url{https://gitlab.com/qumphy/d1-code}.
\section*{Acknowledgments}
The project (22HLT01 QUMPHY) has received funding from the European Partnership on Metrology, co-financed from the European Union’s Horizon Europe Research and Innovation Programme and by the Participating States. Funding for NPL and the University of Cambridge was provided by Innovate UK under
the Horizon Europe Guarantee Extension, grant numbers 10084125 and 10091955. PHC acknowledges support from the British Heart Foundation (grant FS/20/20/34626).

\appendix

\section{UQ methods}\label{sec:UQ_methods}
\subsection{MAP estimation}
\label{sec:MAP_theory}
\subsubsection{Background: Uncertainty from maximum likelihood estimation}\label{sec:MLE}

A popular approach for obtaining uncertainties for neural networks is to assume some parametrised distribution for the predictions and then to use maximum likelihood estimation (MLE) to learn point estimates for these parameters.

In the case of classification, a typical approach is to use a network whose last layer is a softmax to predict class probabilities. A natural distribution to assume for the class probabilities in this case is a categorical distribution (or Bernoulli distribution in the special case of a binary classification). For a neural network classifier with parameters $\pmb{\theta}$ where $p_{ic}(\pmb{\theta})$ denotes the predicted probability output for training sample $i\in\{1,\ldots,N\}$ and class $c\in\{1,\ldots,C\}$, the corresponding likelihood function $p(\pmb{y}|\pmb{\theta})$ for the observed labels $\mathbf{y}=(y_1,\dots,y_N)$ is given by
\begin{equation}\label{eq:class_likelihood}
p(\pmb{y}|\pmb{\theta}):=\prod_{i=1}^N \prod_{c=1}^C p_{ic}(\pmb{\theta})^{\mathbbm{1}(y_i=c)},
\end{equation}
where $y_i\in\{1,\ldots,C\}$ is the training label for the $i^{\textrm{th}}$ training sample and where $\mathbbm{1}$ denotes the indicator function. The negative log likelihood loss function corresponding to (\ref{eq:class_likelihood}) is given by
\begin{equation}\label{eq:class_NLL}-\log p(\pmb{y}|\pmb{\theta})=-\sum_{i=1}^N \sum_{c=1}^C \mathbbm{1}(y_i=c)\log p_{ic}(\pmb{\theta}),
\end{equation}
which is the well-known categorical cross-entropy.

In the case of regression, a common approach is to place a Gaussian assumption on the model outputs and to use maximum likelihood estimation to learn its mean and standard deviation. For a neural network regressor with parameters $\pmb{\theta}$ let $\mu_i(\pmb{\theta})$ and $\sigma_i(\pmb{\theta})$ denote the mean and variance of the output for training sample $i\in\{1,\ldots,N\}$. The corresponding likelihood function $p(\pmb{y}|\pmb{\theta})$ is given by
\begin{equation}\label{eq:regress_likelihood}
p(\pmb{y}|\pmb{\theta}):=\prod_{i=1}^N\frac{1}{\sqrt{2\pi}\sigma_i(\pmb{\theta})}\exp\left\{-\frac{\left[y_i-\mu_i(\pmb{\theta})\right]^2}{2\sigma_i^2(\pmb{\theta})}\right\}.
\end{equation}
The negative log likelihood loss function corresponding to (\ref{eq:regress_likelihood}) is given by
\begin{equation}\label{eq:gnll}
  -\log p(\pmb{y}|\pmb{\theta})=\sum_{i=1}^N \left\{\frac{1}{2}\log\sigma_i^2(\pmb{\theta})+\frac{\left[y_i-\mu_i(\pmb{\theta})\right]^2}{2\sigma_i^2(\pmb{\theta})}\right\}+N\log(\sqrt{2\pi}).
\end{equation}
The uncertainty obtained by this  MLE approach can be thought of as aleatoric uncertainty because it depends upon a point estimate of the distribution parameters and consequently does not take into account the variability of the parameters of the model.

\subsubsection{From MLE to MAP}

The MLE approach described in Section~\ref{sec:MLE} is a frequentist approach, and an alternative is to consider a related Bayesian approach. Taking the classification case first, by Bayes' rule the posterior distribution of the model parameters $p(\pmb{\theta}|\pmb{y})$ is given by
$$p(\pmb{\theta}|\pmb{y})\propto p(\pmb{\theta})p(\pmb{y}|\pmb{\theta}),$$
where $p(\pmb{\theta})$ is the prior distribution for $\pmb{\theta}$. We choose a centred Gaussian distribution $\textbf{N}(0,\eta^2)$ for the prior for each of the $M$ parameters in the network, so that the posterior is given by
$$p(\pmb{\theta}|\pmb{y})\propto\prod_{i=1}^N \prod_{c=1}^C p_{ic}(\pmb{\theta})^{\mathbbm{1}(y_i=c)}\prod_{j=1}^M \frac{1}{\sqrt{2\pi}\eta}\exp\left(-\frac{\theta_j^2}{2\eta^2}\right).$$
The choice of a zero-centred Gaussian is attractive since it leads to a straightforward regularisation term. The parameter $\eta$ must be carefully tuned for a given problem. Ignoring additive constants since they do not affect the optimisation, the maximum a posteriori (MAP) estimate of $\pmb{\theta}$ is then found by minimising the loss function
$$-\sum_{i=1}^N \sum_{c=1}^C \mathbbm{1}(y_i=c)\log p_{ic}(\pmb{\theta})+\sum_{j=1}^M \frac{\theta_j^2}{2\eta^2}.$$

Choosing the same prior distribution for the parameters $\pmb{\theta}$ in the regression case, the MAP estimate of $\pmb{\theta}$ is found by minimising the loss function
$$\sum_{i=1}^N \left\{\frac{1}{2}\log\sigma_i^2(\pmb{\theta})+\frac{\left[y_i-\mu_i(\pmb{\theta})\right]^2}{2\sigma_i^2(\pmb{\theta})}\right\}+\sum_{j=1}^M \frac{\theta_j^2}{2\eta^2}.$$

\subsection{Monte Carlo Dropout}\label{sec:mcd}
Bayesian optimisation is a well-known technique for optimising predictive models that yield predictive distributions for a given input \cite{frazier2018tutorial} from which we can sample. The variance in the sampled predictions enables the estimation of epistemic uncertainty. Given a model $f_{\pmb{\theta}}$ with parameters $\pmb{\theta}$ and a dataset $\mathcal{D}$ composed of $N$ inputs (e.g.\ raw PPG time-series) $x_1, \ldots, x_N \in \mathcal{D}_x$ and ground truth quantities (e.g.\ systolic blood pressure) $y_1, \ldots, y_N \in \mathcal{D}_y$, a prior $p(\pmb{\theta})$ is placed over the model's parameters, where Bayes' theorem is used to acquire a posterior distribution,  
\begin{equation}
    p(\pmb{\theta}|\mathcal{D}) = \frac{p(\mathcal{D}_y|\mathcal{D}_x,\pmb{\theta})p(\pmb{\theta})}{\int_{\pmb{\theta}} p(\mathcal{D}_y|\mathcal{D}_x,\pmb{\theta}')p(\pmb{\theta}')d\pmb{\theta}'}.
    \label{eq:posterior}
\end{equation}
where this particular expression emerges from forcing independence between the inputs and the parameters \cite{jospin2022hands}. In principle, a predictive distribution may be acquired by evaluating, 
\begin{equation}
    p(y|x,\mathcal{D}) = \int_{\pmb{\theta}}p(y|x,\pmb{\theta}')p(\pmb{\theta}'|\mathcal{D})d\pmb{\theta}'.
\end{equation}
However, the denominator (evidence) of Eq. \ref{eq:posterior} is intractable to evaluate given the large number of parameters of typical deep learning models \cite{gal2016dropout}. This motivates a need for more efficient techniques.

Variational inference is one approach \cite{gal2016dropout}, where instead the evidence may be approximated with a more tractable variational distribution $q(\pmb{\theta}')$, which is acquired by optimising the evidence lower bound (ELBO), 

\begin{equation}
    \text{KL}(q(\pmb{\theta}'|\pmb{\theta})||p(\pmb{\theta})) - \mathbb{E}_{q(\pmb{\theta},\pmb{\theta}')}\log(p(\pmb{\theta},\pmb{\theta}')),
\end{equation}
where $\pmb{\theta}'$ are values that parametrise the variational distribution (e.g.\ $\pmb{\theta}'=(\pmb\mu,\pmb\sigma^2)$ where $\mu$ is the mean and $\sigma^2$ the variance of a Gaussian), and KL is the Kullback-Liebler Divergence.

However, the computational expense of variational inference is still high \cite{gal2016dropout}. Therefore, further efforts have been placed towards developing even more efficient and effective techniques.

Monte Carlo Dropout (MCD) \cite{gal2016dropout} is one such approach that leverages the fact that training a model with dropout applied throughout the architecture approximates variational inference. In the most generic implementation, dropout is left active for evaluation, i.e.\ the weights and biases of the relevant neural network layers are set to 0 with probability $\omega$. Then, the output is estimated through $T$ iterations of the stochastic forward pass for each input. The resulting set of predictions provides an estimate of model uncertainty. Variants of this approach additionally model homoscedastic or heteroscedastic aleatoric uncertainty \cite{kendall2017uncertainties,gal2016dropout}. We model heteroscedastic aleatoric uncertainty for both tasks.

For classification, the stochastic model outputs means $\mu_{i,c,t}$, and variances $\sigma_{i,c,t}^2$ are used to parametrise a Gaussian distribution for the logits $z_{i,c,t}$ for each class $c$, input $i$ and stochastic forward pass iteration $t$. The means and variances are independently evaluated for $T$ iterations. For each iteration $t$, we draw $K=100$ samples from the distribution $z_{i,c,t,k}\sim\mathcal{N}(\mu_{i,c,t},\sigma_{i,c,t}^2)$ of the logits of each class and apply a softmax to each sampled logits vector $\mathbf{z}_{i,t,k}$ to get a vector of class probabilities $\mathbf{p}_{i,t,k}$. We disentangle aleatoric from total uncertainty using a custom averaging procedure (inspired by Kendall et al.\ \cite{kendall2017uncertainties}), shown in Algorithm 2. The results are then aggregated, where the variance of the predictions widens the distribution. Uncertainty is expressed as an entropy of the relevant distributions predicted from the sampling procedure, as shown in Algorithm 2.

\begin{algorithm}

\caption{Monte Carlo Dropout Classification Training Loop}
\begin{algorithmic}
\label{alg:train_class}
\FOR{each batch}
\STATE Compute predicted mean $\mu_{i,c}$ and variance $\sigma^2_{i,c}$ for each class $c$ for each input $i$ in the batch
\FOR{$k = 1$ to $K$}
\FOR{each class $c$}
\STATE Sample $\epsilon_{i,c,k} \sim \mathcal{N}(0,1)$
\STATE Compute sampled logit $z_{i,c,k} = \mu_{i,c} + \sigma_{i,c}\epsilon_{i,c,k}$
\ENDFOR
\STATE Compute $\textbf{p}_{i,k} = \text{softmax}(\textbf{z}_{i,k})$
\ENDFOR
\STATE Compute average $\bar{\textbf{p}_i} = \frac{1}{K}\sum_{k=1}^{K}\textbf{p}_{i,k}$
\STATE Evaluate NLL on $P$ where $P$ is the batch of $\bar{\textbf{p}}_i$ values.

\ENDFOR
\end{algorithmic}
\end{algorithm}

\begin{algorithm}
\label{alg:eval_class}
\caption{Monte Carlo Dropout Classification Evaluation Loop}
\begin{algorithmic}
\FOR{each input}
\FOR{$t=1$ to $T$}
\STATE Acquire the predicted mean $\mu_{c,t}$ and variance $\sigma^2_{c,t}$ for each class $c$
\FOR{$k = 1$ to $K$}
\FOR{each class $c$}
\STATE Sample $\epsilon_{c,t,k} \sim \mathcal{N}(0,1)$
\STATE Compute sampled logit $z_{c,t,k} = \mu_{c,t} + \sigma_{c,t}\epsilon_{c,t,k}$
\ENDFOR
\STATE Compute $\textbf{p}_{t,k} = \text{softmax}(\textbf{x}_{t,k})$
\ENDFOR
\STATE Compute average $\bar{\textbf{p}}_{t} = \frac{1}{K}\sum_{k=1}^{K}\textbf{p}_{t,k}$
\ENDFOR
\STATE $H_{\text{ale}} = \frac{1}{T}\sum^T_{t=1}H(\bar{\textbf{p}}_{t})$
\STATE $H_{\text{total}} = H(\frac{1}{T}\sum^T_{t=1}\bar{\textbf{p}}_{t})$ \\
Where $H_{\text{ale}}$ is the aleatoric uncertainty expressed as an entropy.
\ENDFOR
\end{algorithmic}
\end{algorithm}

For regression, we take the approach of Kendall et al.\ \cite{kendall2017uncertainties}. The stochastic model $f_{\pmb{\theta}, \omega}(x_i)$ outputs the mean and variance of a Gaussian distribution for each predicted quantity. Specifically, via MAP estimation, the model outputs the means $\mu_{SBP,i, t}$, $\mu_{DBP,i,t}$, and variances $\sigma_{SBP,i,t}^2$, $\sigma_{DBP,i,t}^2$ for a given input $x_i$ (e.g.\ a raw PPG time-series) that parametrise independent Gaussian distributions for each stochastic forward pass $t$ of the model for a total of $T$ iterations. We then estimate the mean, and aleatoric and epistemic uncertainties using of the Law of Total Variances \cite{valdenegro2022deeper} as shown in Algorithm 3.

\begin{algorithm}
 \label{alg:eval_regression}
\caption{Monte Carlo Dropout Regression Evaluation Loop}
\begin{algorithmic}
\FOR{each input $i$}
\FOR{$t=1$ to $T$}
\STATE $\mu_{t}, \sigma^2_{t} = f(\pmb\theta)$ 
\ENDFOR
\STATE $\mu_i = \frac{1}{T}\sum_{t=1}^T\mu_{i,t}$
\STATE $\sigma_{\text{epi}, i}^2= \frac{1}{T}\sum^{T}_{t=1}(\mu_{i,t} - \frac{1}{T}\sum_{t=1}^{T}\mu_{i,t})^2$
\STATE $\sigma_{\text{ale}, i}^2= \frac{1}{T}\sum_{t=1}^{T}\sigma_{i,k}^2$
\ENDFOR
\end{algorithmic}
\end{algorithm}

\subsubsection{Uncertainty disentanglement}
Recent work has shown that uncertainties disentangled using the Law of Total Variances \cite{valdenegro2022deeper} or other popular techniques produce aleatoric and epistemic uncertainty estimates that are significantly correlated \cite{mucsanyi2024benchmarking}. While some correlation is inevitable given higher aleatoric uncertainty examples are less-well represented in the training data (resulting in higher epistemic uncertainty), this correlation should be low. It is advisable to keep these limitations in mind when interpreting disentangled uncertainty estimates to inform dataset curation, model selection, and diagnosis. While we do not further address these concerns here, this will be the subject of future work.  

\subsection{Deep Ensembles}
With deep ensembles \cite{lakshminarayanan_simple_2017}, we capture model variability by training several instances of the same architecture, on the same task, using the same dataset, where each model is initialised with different parameters and trained independently. While it is heuristic in its formulation, it has a Bayesian interpretation \cite{fort2019deep}, where each member of the ensemble can be thought of as a sample from the posterior distribution of the outputs. It has been hypothesised that the superior quality of uncertainty estimates often reported relative to other approaches may be attributed to the fact that this approach enables a broader sampling of the posterior \cite{fort2019deep}. For example, with MCD, each dropout model is derived from the same `parent' model that may focus on just one mode, whereas with DE, the less restrictive form of sampling may cover more modes of the posterior.

We employ variants of Algorithms 1, 2 and 3 for training and  evaluation, where $T=50$ dropout samples are replaced with $T=5$ ensemble members. As with MCD, we retain the use of $K=100$ samples for noise corruption of the logits in classification. Aleatoric uncertainty is modelled using likelihood-based losses (see Tables  
\ref{tab:train_details_af_tasks},\ref{tab:train_details_bp_tasks}).

\subsection{Quantile Regression}
The aim of quantile regression \cite{steinwart2011estimating} is to predict the values corresponding to a set of quantiles. Here, we implement the pinball loss as described in (\ref{eq:quantile}).

\subsection{Post-hoc calibration methods}

Post-hoc calibration methods are commonly used mainly because they do not require retraining of the models to adjust their predictions to improve the reliability of uncertainties. These techniques can be broadly categorised into scaling-based, binning-based and distribution-based approaches. Scaling techniques (Temperature Scaling (TS) \cite{temperature_scaling}, Logistic Calibration \cite{wang2023calibration}) apply a parametric transformation to the model's logits (in the case of classification) while binning techniques partition predictions into bins and calibrate them based on the observed frequency of correct predictions within each bin (Histogram Binning (HB) \cite{gupta2021distribution}, Bayesian Binning into Quantiles (BBQ) \cite{naeini2015obtaining}). 
Distribution-based approaches (Conformal Prediction \cite{angelopoulos2021gentle}) adjust prediction distributions to provide well-calibrated uncertainty estimates.

\subsubsection{Temperature Scaling}
\label{subsubsection_Temperature Scaling}

Temperature scaling (TS) \cite{temperature_scaling} is a post-processing calibration technique commonly applied to deep learning models to improve the reliability of their uncertainties. 

For classification, the uncertainty is given by the output probabilities (i.e.\ the confidences) of the model. These are adjusted to better reflect the true likelihood of each predicted class. Since the technique operates on the predicted logits rather than the features or parameters of the model, it has the benefit not to alter the model’s accuracy. Indeed, TS preserves the relative ranking of the probabilities, ensuring that the model's classification accuracy remains unchanged. Therefore, it is a popular choice for scenarios where calibrated probabilities are essential, such as in medical diagnosis, autonomous driving, and other fields where understanding model uncertainty is critical. Moreover, it is straightforward and computationally efficient.
A scalar parameter, denoted \( T > 0 \) (known as the temperature), is introduced and applied to the logits output by a trained model before passing them through a softmax function to produce calibrated probabilities. Given an uncalibrated model’s logits \(\mathbf{z}\), where \(\mathbf{z} = [z_1, z_2, \dots, z_C]\) where \( C \) is the number of classes, the calibrated probabilities are computed as:

\[
p_i = \frac{\exp(z_i / T)}{\sum_{j=1}^C \exp(z_j / T)},\quad i=1,\ldots,C
\]

where \( p_i \) is the probability of class \( i \) after temperature scaling.
\( T \) is optimised to minimise a calibration objective, typically the negative log-likelihood (NLL) on a validation set. This effectively “softens” the logits, as higher values of \( T \) yield a more uniform distribution, reducing the model's overconfidence in its predictions. \( T \) can be optimised using gradient descent on the NLL, without the need to retrain the model itself. 
Once \( T \) is determined, it can be applied in any setting where the model’s calibrated probabilities are desired. The main limitation of the TS approach is that it assumes that a single scalar can be used to globally calibrate all output probabilities, which may not be sufficient for models presenting more complex miscalibration.

For regression, the uncertainty is given by a variance parameter. Temperature scaling in this context is referred to as variance scaling \cite{variance_scaling} where the uncalibrated variance is multiplied by a scaling factor known analytically for Gaussian distributions.

\subsubsection{Isotonic Regression}
\label{subsubsection_Isotonic Regression}
Isotonic regression (IR)~\cite{isotonic_regression_quantile_calibration}, initially proposed by~\cite{zadrozny2002transforming}, is a non-parametric, piecewise-constant regression method that is useful when the relationship between the inputs and the outputs is assumed to be monotonic. 
Given input-output pairs $\left(x_i, y_i\right)_{i = 1, \ldots, n}$,  IR finds the mapping function $g$ that minimises the sum of squared errors between the observed values $y_i$ and the fitted values $g(x_i)$ assuming monotonicity: 
\begin{eqnarray}
\min_f &\sum_{i=1}^n \left(y_i - g(x_i)\right)^2    \nonumber \\
\mbox{subject to }&g(x_i) \leq g(x_j), \quad \forall x_i < x_j \nonumber
\end{eqnarray}
The method does not assume any functional form for $g$ making it very flexible despite the loss of model interpretability since it gives only a piecewise-constant function with no equation compared to parametric approaches. 

In the context of binary classification ~\cite{isotonic_regression}, the objective is to find the mapping function $g$ between the classifier's output probabilities and their calibrated counterparts. Given the classifier scores, $z$, and the true labels, $y$, i.e.\ the pair $\left(z_i, y_i\right)_{i=1, \ldots, n}$ the objective is to find $g$ such that $g(z) \simeq \mathbf{P}(y=1|z)$ through optimisation. Such monotonicity transformation is helpful to interpret model outputs as probability estimates despite the lack of equations to describe $g$.
To extend IR for use in a multiclass classification problem, independent isotonic regression models are created, one per class using a one-vs-all strategy. Given $C$ classes, $C$ models are generated and the true labels become a binary indicator $y_{i, c}$ equal to 1 only when the sample belongs to class $c$ and 0 otherwise. This leads to estimate $C$ isotonic functions $\left(g_{c}\right)_{c=1, \ldots, C}$.
An extra normalisation step across all classes is required to ensure that calibrated probabilities sum to 1.

\subsection{Conformal prediction}\label{appendix:conformal}

Conformal prediction is a UQ technique for classification and regression with statistical guarantees, irrespective of the accuracy of the underlying model or the underlying data distribution, see \cite{angelopoulos2021gentle} for a pedagogical introduction. For classification, it outputs a prediction set—that is, a set of labels—guaranteed to contain the true label with a user‐specified confidence level. For regression, the method outputs prediction intervals whose widths adjust to the local noise or heteroscedasticity of the data. This is done by calculating nonconformity scores (e.g.\ the negative log‐probability of the true class or the absolute residual) on a held‐out calibration set and then taking an appropriate quantile (several may be considered to consider various regions of the distribution) of the scores as a threshold. In the following, we describe the specifics of conformal prediction approaches used in this work, distinguishing between regression and classification.

\subsubsection{Regression}\label{appendix:conformal_reg}
In the regression case, the procedure depends if a model provides parametric (Gaussian) output distributions from training with a Gaussian Negative Log Likelihood Loss or outputs specific quantiles as per training with quantile loss as the loss function. In the former case, conformal prediction can be used to statically calibrate the predicted standard deviation per data point, see \cite{angelopoulos2021gentle}. In the latter case, we rely on the framework of conformalised quantile regression \cite{romano2019conformalized}, which starts from heuristic predicted quantiles and turns them into conformalised predicted quantiles that fulfil a coverage guarantee in the sense that newly formed prediction intervals contain the actual value with a predefined confidence level.

In our implementation, we focus on the latter approach, conformalising quantiles that are either heuristically predicted using the QR UQ technique (denoted in Tables~\ref{table:alexnet_calib_results}-\ref{table:resnet_calibfree_results} as CQR), or derived from a Gaussian distribution assumption from training a model with a GNLL loss (denoted in Tables~\ref{table:alexnet_calib_results}-\ref{table:resnet_calibfree_results} as CMAP). In both cases, the same nonconformity score from \cite{romano2019conformalized} is used to enable a comparison between conformal approaches applied to different underlying UQ techniques.

\subsubsection{Classification} \label{appendix:conformal_classification}
In the classification case, one option is to apply the conventional split conformal prediction (SCP) approach to provide prediction sets with coverage guarantees, see \cite{angelopoulos2021gentle}, i.e.\ sets of predicted classes that cover the correct class with a certain predefined confidence level. However, these methods are conventionally used in the context of a large number of classes and are difficult to interpret in the case of binary prediction problems, as considered in this work, see also \cite{Pereira2020}.

A more appropriate choice in this case are other conformal prediction variants, namely Venn-ABERS predictors \cite{vovk2015large}. They start from predicted output probabilities and turn them into calibrated output probabilities by fitting two isotonic regression models for each data point (one for each possible label in the binary case). This leads to prediction intervals for the output probabilities, which again fulfil statistical guarantees.

\section{Uncertainty evaluation}
\label{sec:uq_eval}
\subsection{Local vs.\ global reliability}
\label{sec:local_global}
An assessment of reliability should cater to how predictions of uncertainty will be used in practice. For example, if a single prediction is to be used to inform diagnosis, then the corresponding single uncertainty estimate should be well-calibrated (known as individual reliability) \cite{pernot2023calibration}. However, this contrasts with the majority of reliability metrics which tend to assess whether the average uncertainty over a subset of examples is correlated with their corresponding average error/accuracy (known as local reliability). Some metrics also evaluate reliability over the whole test set (global reliability). Local reliability metrics often bin predictions by the magnitude of the predicted uncertainty (sometimes referred to as size-stratified metrics \cite{angelopoulos2021gentle}).

\subsection{Adaptivity}
Depending on how the models are used to inform diagnosis, it may also be advantageous to first separate the data into distinct subsets based on something \textit{other than} the magnitude of predicted uncertainties, and then subsequently apply local reliability metrics to each set individually. For example, for AF classification, it may be of interest to acquire information about reliability quality for each class individually. For regression, one could theoretically observe how reliability varies with signal quality. Computing the reliability metrics for different subsets of data enables an assessment of a model's adaptivity \cite{pernot2023calibration}.

\subsection{Calibration-sharpness paradigm}
The use of local calibration techniques for regression models is motivated by the need to go beyond global calibration and address small-scale local reliability. An alternative approach for doing this is the so-called calibration-sharpness paradigm~\cite{gneiting2007probabilistic}.

While calibration captures the extent to which the estimated uncertainties correlate with prediction error, sharpness captures the extent to which confidence intervals concentrate around the predicted value. 

It has been argued that both calibration and sharpness are important in validating uncertainty~\cite{gneiting2007probabilistic}. Ideally we would like our model's uncertainties to be as trustworthy and as informative as possible. It is possible to have a trustworthy but overly-defensive model which is well-calibrated but unsharp: for example a model which assigns the same (well-calibrated) probability to every data sample (e.g.\ 50\% confidence for a binary classification task with equal populations for each class). On the other hand, it is possible to have a model whose predicted uncertainties are too small (tight), thereby underestimating the true variability. In the calibration-sharpness paradigm, the aim is to optimise sharpness subject to constraints on calibration.

\subsection{Expressions of uncertainty and comparisons between them}
For regression, uncertainty is often expressed as a predicted variance for each measurand (predicted quantity, e.g.\ SBP and DBP for BP regression), or as a prediction interval (e.g.\ upper and lower quantiles). The reliability metrics implemented here for regression require an assumed form of the distribution (aside from PICP). For UQ methods that do not inherently assume a form, we choose a Gaussian. For all other methods that require an assumed form \textit{a priori}, we also choose a Gaussian. This provides a convenient means to compare between approaches that output uncertainties in either form of expression.

For regression tasks, the first widely used calibration metrics catered to the interval expression of uncertainty. These assessed whether the frequency with which the ground truths fall below a chosen quantile of their respective predicted distribution matches the numerical value of the quantile. These are known as coverage-based metrics. Coverage reliability metrics are often used to assess global calibration, though one can apply each to subsets of test predictions to enable an assessment of small-scale calibration. For the case where variance is used to express uncertainty \cite{levi2022evaluating}, one considers whether the magnitude of the estimated standard deviation correlates with the magnitude of the prediction error. The variance-based metric used here (ENCE) is inherently local. 

For classification, uncertainty may be expressed as the probability of the predicted class, or the entropy of the predicted probability distribution. For binary classification, there is a nonlinear correspondence between the two, where the Expected Calibration Error (ECE) is used for class probabilities, and the Uncertainty Calibration Error (UCE) is used for entropy (see Table \ref{tab:uncertainty_metrics}). Entropy is generally preferred for encoding uncertainty in the multi-class case, as it captures how the probabilities are spread across the whole distribution of classes.

\subsection{Binning}
The binning strategy employed when computing local/adaptive reliability metrics can have a significant effect on the value of the metric. For example, \cite{pernot2023properties} mentions that the value of the ENCE (described in Appendix \ref{sec:ence}) can scale with the number of bins. For classification, metrics like the smECE, introduced in Table \ref{tab:uncertainty_metrics}, attempt to mitigate the issues surrounding binning strategies, through kernel density estimation. We include metrics that employ a range of binning strategies (e.g.\ ACE) to highlight its impact on the evaluation of uncertainty reliability.

\section{UQ Evaluation Metrics}
\label{sec:appendix_eval_metrics}

\subsection{Metrics for classification}\label{conditional_classification}

We first describe some uncertainty reliability metrics for predicted uncertainties acquired from classifiers in which the default implementation is local (i.e.\ binned by the magnitudes of estimated uncertainties). For all metrics that bin based on the magnitude of the model's uncertainty, we use a \textbf{consistent number of bins of 15}. The following reliability metrics are described in the multi-class case, and any adaptations needed for these metrics to be applied in the binary classification case are discussed separately in each description. 

Some metrics, such as Expected Calibration Error (ECE), capture the uncertainty of model outputs in terms of class probabilities. Others, such as Uncertainty Calibration Error (UCE), capture the uncertainty of model outputs using entropy. 

Predicted uncertainties are typically compared with observed class proportions (e.g.\  fraction of examples with a correct prediction), and with this in mind capturing uncertainty in terms of class probabilities is a natural approach. On the other hand, capturing uncertainties in terms of entropy extends better to the multi-class case (e.g.\ output probabilities of [.1, .1, .1, .1, .1, .95] are more uncertain than [0, 0, 0, 0, .5, .95] and entropy captures this). For this reason we consider both approaches for capturing uncertainty, which in addition allows us to study the interplay between `calibrated entropy' and `calibrated class probabilities'.

\subsubsection{Expected Calibration Error (ECE)}
The Expected Calibration Error (ECE) \cite{guo2017calibration} calculates a weighted average of the absolute difference between the model's accuracy and confidence. This is achieved by assessing the calibration of the model predictions by binning the predicted confidences (the maximum class probability) in equal-width bins, and determining the difference between the fraction of correct predictions in the bin (accuracy) and the average of the confidences in the bin. The ECE calculates a weighted average of these local calibration errors across all bins, and is given by 

\begin{equation}
    \text{ECE} = \sum_{m=1}^{M} \frac{|B_{m}|}{N} |\text{acc}(B_{m}) - \text{conf}(B_{m})|, 
\end{equation}

where $M$ is the chosen number of bins, $N$ the total number of predictions, $|B_{m}|$ the size of bin $m$, and $\text{acc}(B_{m})$ and $\text{conf}(B_{m})$ are the accuracy and average confidence within bin $m$ respectively. The ECE scalar lies within the interval [0, 1], and allows for comparison of calibration techniques for different models. 

The ECE suffers from the same drawback as all `average' calibration methods, namely that it is unable to detect variation in calibration in different subsets of the data. For example, it was shown in~\cite{lavesuncertainty} that it can be minimised by a model which constantly predicts the marginal distribution of the highest-probability class. Since it is based on the maximum probability class and not entropy, ECE also does not capture the behaviour of the probabilities in the other classes. The ACE and UCE metrics, on the other hand, provide a way of capturing the probabilities of both the predicted class and the non-predicted classes.

In addition, due to models being typically overconfident, model outputs exhibit a skew towards the right-hand side of reliability diagrams, which results in a small proportion of bins contributing significantly to the ECE. Whilst choosing a larger number of bins can reduce the bias, this increases the variance of the discrepancy in the uncertainty magnitude and prediction error measurement per bin, as more bins become sparsely populated. This can be addressed by redefining the binning scheme, as done for the Adaptive Calibration Error (ACE) metric \cite{nixon2019measuring}.

\subsubsection{Adaptive Calibration Error (ACE)}

The ACE modifies the ECE by redefining the binning scheme from equal-width to equal-frequency bins \cite{nixon2019measuring}. This aims to address the high variance concern with equal-width bins, where the large bin number reduces the bias but also increases the variance of each metric measurement per bin. With equal-frequency bins, this high variance is reduced, motivated by the desire to address regions with a greater number of predictions to get a better estimate of the calibration error. 

In the general multi-class classification case, for $K$ classes with a chosen number of ranges $R$, the ACE is given by 

\begin{equation}
    \text{ACE} = \frac{1}{KR} \sum_{k=1}^{K} \sum_{r=1}^{R} |\text{acc}(B_{r, k}) - \text{conf}(B_{r, k})|,
\end{equation}

where $\text{acc}(B_{r, k})$ and $\text{conf}(B_{r, k})$ are the accuracy and confidence for class $k$ in range $r$ respectively. $B$ here stands for the samples in a bin.

ACE is more sensitive to the chosen number of ranges $R$, with ranges becoming very large if there are sparsely populated regions in the model predictions' distribution, and the number of bins becomes $K \cdot R$ for the multi-class setting. The sensitivity to the chosen number of ranges can be reduced with a more adaptive binning scheme, in which the bins are not simply equal-width or equal-frequency. Such adaptive binning schemes lose interpretability across different models however, as the range boundaries are not consistent across predictions from models with different UQ methods. This motivates the implementation of a metric that characterises the model uncertainty in a consistent way across different models. 

It is important to consider the impact of infinitesimal softmax outputs on calibration error calculations \cite{nixon2019measuring}. Such values introduce large biases in sparsely populated bins, including the last bin that will contain the extreme confidence values. ACE can be thought of as a special case of the thresholded adaptive calibration error (TACE) \cite{nixon2019measuring} with a threshold value set to 0. Choosing a relevant threshold can help avoid large biases in the calibration error metric.

\subsubsection{Uncertainty Calibration Error (UCE)}\label{sec:UCE}

The Uncertainty Calibration Error (UCE) uses an equal-width binning approach similar to ECE, however within each bin the difference between the model \textbf{in}accuracy and the average normalised entropy is calculated~\cite{lavesuncertainty}. These differences are weighted by the empirical probability of finding the entropy values within the current bin, $\frac{|B_{m}|}{N}$, where $|B_{m}|$ is the size of bin $m$, and $N$ is the total number of data points. 

Entropy is the chosen measure of the model's prediction uncertainty, and is given by

\begin{equation}
    \mathcal{H}(\mathbf{p})=-\sum_{k=1}^{K} p_{k} \log p_{k}
\end{equation}

where $p_{k}$ is the prediction probability for class $k$ out of total classes $K$. However, the entropy scales with the number of classes $K$, preventing comparison of model uncertainty or model calibration across different data sets. The normalised entropy is used instead for interpretability, and is given by

\begin{equation}
    \Tilde{\mathcal{H}}(\mathbf{p}) = -\frac{1}{\log K} \sum_{k=1}^{K} p_{k} \log p_{k}.
\end{equation}

Perfect calibration as determined by the UCE corresponds to a normalised entropy that is equal to the misclassification rate. As is the case for the ECE metric, an approximation of the UCE is calculated using a binning scheme. The model normalised entropy values are assigned to a chosen number of equal-width bins, as follows:

\begin{equation}
    \text{UCE} = \sum_{m=1}^{M} \frac{|B_{m}|}{N} |\text{err}(B_{m}) - \text{uncert}(B_{m})|,
    \label{eq8}
\end{equation}

where $\text{err}(B_{m})$ is the average inaccuracy for all samples in bin $m$, and $\text{uncert}(B_{m})$ is the average normalised entropy for all samples in bin $m$. The UCE is non-zero for a model that only predicts the marginal distribution of the predicted class, and is less sensitive to the total number of bins \cite{lavesuncertainty}, thereby allowing for more consistent comparison of calibration across different models trained on the same data (this is best illustrated in Figure 1 of \cite{lavesuncertainty}, where UCE values for models trained with different Bayesian UQ methods have consistent UCE for varying bin number, whereas ACE scales with bin number and the ranking of model performance is bin number dependent). 

For the binary classification case, a slope of 0.5 indicates perfect calibration, as a maximally uncertain prediction corresponds to a normalised entropy of 1, which therefore has a prediction accuracy of 0.5. In order to account for this relationship, $\text{uncert} (B_{m})$ in equation (\ref{eq8}) is halved. 

\subsubsection{Variation Calibration Error}
While some theoretical justification for the UCE is given in~\cite{lavesuncertainty} in the limit of infinitely many classes, it is not clear why we should expect linear correlation between predicted entropy and misclassification rate. An alternative approach with stronger theoretical justification was recently given in~\cite{thompson2025variation}, where $\text{err}(B_{m})$ is replaced by the entropy of the binary probability distribution defined by the observed correct and incorrect classification proportions respectively. It is shown in~\cite{thompson2025variation} that, in the general multi-class case, this approach in fact defines a family of calibration metrics in which the ranked probability distributions of predictions and observations are compared according to some measure of variation, referred to as the \textbf{Variation Calibration Error (VCE)}. Different measures of variation are possible; see~\cite{bilson2025metrological} for further examples. If the measure of variation is taken to be the maximum probability (confidence), the usual notion of confidence calibration is recovered.

\subsection{Global classification metrics}\label{average_classification}

We next describe metrics which assess the predicted uncertainties from classifiers in the average sense, where the model uncertainties are not binned conditional on their magnitude. 

\subsubsection{Smooth Expected Calibration Error (smECE)}\label{sec:smECE}
Motivated by the pathologies of the binning-based ECE, \cite{blasiok2023smooth} we introduce the Smooth ECE (smECE). They employ the use of reflected Gaussian kernels to provide a kernel density estimate of the empirical distribution of the residuals $r_{i}:=y_{i}-f_{i}$ between a model's prediction $f_{i} \in [0,1]$ and the ground truth class $y_{i} \in {0,1}$, for $i=1, 2, ..., N$ examples. The usage of a reflected kernel accounts for boundary effects of the $[0,1]$ interval. The details  of the kernel smoothing method is beyond the scope of this paper, but important kernel parameters, e.g.\ the kernel choice and bandwidth, are chosen automatically in a theoretically justified way. The method determines an ideal kernel bandwidth that is proportional to the calibration error. In this case, the smECE is defined as the integral between the kernel smoothed residuals between model predictions and ground truths, and the diagonal representing perfect calibration. The smECE metric is implemented using the \textsc{relplot} Python package \cite{blasiok2024smooth}. 

Whilst the smECE provides a solution to the issues associated with binning, as discussed in Section \ref{other}, it stills retains some of the constraints of the ECE, notably that it does not account for model probabilities of all classes. In addition, smECE indicates good calibration even when the model constantly predicts the marginal distribution of the highest-probability class \cite{lavesuncertainty}. 

\subsubsection{Negative Log Likelihood (NLL)}\label{sec:NLL}

The negative log-likelihood (NLL) is a common metric used to assess model performance in classification tasks. It is a proper scoring rule which, as discussed above, is minimised if and only if the predicted distribution is equivalent to the true distribution. 

In the binary classification setting, the NLL is given by:

\begin{equation}
    \text{NLL} = - \sum_{i=1}^{N} [y_{i}\log(p(y_{i})) + (1-y_{i})\log(1-p(y_{i}))],
\end{equation}

where $y_{i}$ represents the ground truth class label, and $p(y_{i})$ is the model's probability for class $y_{i}$. As discussed earlier, a proper scoring rule considers both calibration and sharpness. The NLL is commonly used as an optimisation objective for training neural networks, and can be optimised by overconfident predictions as the NLL penalises low confidence scores assigned to the correct class, and high confidence scores assigned to the incorrect classes \cite{ashukha2020pitfalls}. This can lead to uncalibrated, overconfident model predictions that yield lower NLL metric results. Despite these potential drawbacks, due to its nature as a proper scoring rule and classification optimisation objective, the NLL is a common metric used to assess the reliability of uncertainties, and we present model results for the NLL to allow comparison to previous work.

\subsection{Global regression metrics}\label{sec:average regression}
Here, we discuss metrics that evaluate the quality of uncertainties over the whole population of test examples. 

\subsubsection{Coverage-based metrics}
A model trained with a Gaussian NLL loss will predict a mean and variance that can be used to parametrise a Gaussian distribution. For each test input, one computes the cumulative prediction function, and then subsequently the confidence interval for a given confidence level. One then records the frequency with which the ground truths lie within their respective confidence intervals. This is evaluated for a range of confidence levels~\cite{kuleshov2018accurate}. A well calibrated model should have the ground truths fall within their respective X\% confidence intervals X\% of the time: $\frac{1}{N}\sum_{n=1}^{N} \mathbb{I}\{y_n \leq F_{n}^{-1}(p)\} = p \text{ for all } p \in [0,1] $, where $F^{-1}$ is the quantile function of the predicted distribution evaluated at the confidence level $p$, and $y_n$ is the corresponding ground truth. 

For models or post-hoc UQ techniques that do not output a known distribution, such as conformal prediction or quantile regression, coverage can still be assessed in a similar way, just at the specified confidence level e.g.\ 2 standard deviations from the mean ($\sim$95\% for Gaussian). This is known as the Prediction Interval Coverage Probability (PICP). The PICP metrics are presented as ratios of achieved coverage to target coverage values, corresponding to $0.6826$ and $0.9544$ for the $1$ and $2$ standard deviation confidence intervals respectively. An optimal PICP result is thus centred on $1$.

If multiple quantiles can be evaluated, then a calibration curve can be plotted showcasing the observed frequency at which the ground truths lies within each chosen confidence interval. The quoted Coverage Calibration Error (CCE) metric for a series of predicted quantiles is calculated as

\begin{equation}
    \text{CCE} = \sum_{j=1}^{M} w_{j} \cdot (p_{j}-\hat{p}_{j})^{2},
\end{equation}

where $w_{j}$ are weights, $M$ is the number of confidence levels, and $\hat{p}_{j}$ is the empirical frequency (or coverage) of the ground truth $y_{t}$ falling below the confidence level $p_{j}$~\cite{kuleshov2018accurate}:

\begin{equation}
    \hat{p}_{j} = \frac{\left| \left\{y_{t} | F_{t}(y_{t}) \le p_{j}, t= 1,\ldots,T\right\}\right|}{T}.
\end{equation}

For our case, we choose all $w_{j}=1$, as done in the original paper. 
  
It is known that perfect calibration scores can be achieved even when the predictions and ground truths are statistically independent. Despite these weaknesses, the metric could be useful in the case where the data has significant outliers as it is count-based, and does not scale with the magnitude of the difference in predicted uncertainty and prediction error.

\subsubsection{Continuous Ranked Probability Score (CRPS)}

One way to assess the sharpness of the distribution is by the Mean Prediction Interval Width (MPIW), which is the interval width for a given confidence. For example, for a Gaussian distribution and a confidence of 68\% the MPIW is just the mean standard deviation. For more complex distributions, e.g.\ bimodal distributions, a little bit of care is necessary. One solution might be to take the sum of the interval widths of the intervals with the highest probability mass. The goal is to minimise this score.

The Continuous Ranked Probability Score (CRPS) is similar to the confidence interval-based metric, but instead of assessing how well the predicted distribution captures the ground truth for a range of quantiles, it does so over the whole distribution. Here, a model trained via a Gaussian NLL loss will output a distribution, expressed as a CDF. We convert the ground truth into a degenerate distribution (Heaviside step-function), and compare the squared difference between this and the predicted distribution CDF:

\begin{equation}
    \text{CRPS}(F, y) = \int_{-\infty}^{\infty} [F(x) - \mathbbm{1}_{x\leq y}]^{2}dx, 
\end{equation}

where $F(x)$ is the CDF of the predicted distribution, and $\mathbbm{1}$ is the indicator function representing the degenerate distribution of the ground truth observation $y$. 

If the model prediction distribution is a univariate Gaussian, there exists an analytical expression for the CRPS:

\begin{equation}
    \text{CRPS}( \mathcal{N}(\mu, \sigma^2), y) = 
    \sigma\left\{2\phi\left(\frac{y-\mu}{\sigma}\right) + 
    \left(\frac{y-\mu}{\sigma}\right)\left[2\Phi\left(\frac{y-\mu}{\sigma}\right)-1\right] - \frac{1}{\sqrt{\pi}}\right\},
\end{equation}

where $(\mu, \sigma^{2})$ are the mean and variance of the output Gaussian distribution, and $\phi$ and $\Phi$ are the PDF and CDF of the standard Gaussian respectively \cite{grimit2006continuous}. The CRPS allows us to compare the outputs of UQ methods that give a Gaussian distribution (Deep Ensembles or Monte Carlo Dropout) to conformal prediction, through the conversion methods outlined in \ref{sec:interval_conversion}.

The CRPS is a proper scoring metric, which means that it captures both calibration and sharpness.

\subsection{Local calibration metrics for regression}\label{sec:local regression}
A model's predicted uncertainties are only useful if they reflect the doubt of the model's prediction for their respective input (i.e.\ are `individually' calibrated \cite{pernot2023calibration}). However, average calibration metrics computed over the whole test set do not directly assess this. For example, with the confidence interval-based metrics, calibration is assessed by whether the observed frequencies from the whole population of test examples match the confidence level used to compute the intervals. The general approach that `local' metrics take to address this limitation is to bin the data according to some parameter, for example the model uncertainties. While binned estimates are still not perfectly local, they provide a means to understand how calibration may vary over the whole population of test examples.

\subsubsection{Expected Normalised Calibration Error (ENCE)}\label{sec:ence}
The expected normalised calibration error (ENCE) is an extension of the ECE to the regression case. As for binning-based metrics in Section~\ref{conditional_classification}, in our implementation of the ENCE we use a number of bins of 15 for consistency across the different methods. The ENCE assesses the difference between the mean estimated variance (MV) and the mean squared error (MSE), weighted across the number of samples. Using a binning approach, where the model uncertainties are sorted in increasing order and then binned into equal-size intervals, the ENCE is given by

\begin{equation}
    \text{ENCE} = \frac{1}{M} \sum_{i=1}^{M} \frac{|\text{MV}_{i}^{1/2} - \text{MSE}_{i}^{1/2}|}{\text{MV}_{i}^{1/2}}
\end{equation}

where the absolute difference between the root mean squared error (RMSE) and the root mean variance (RMV) per bin is normalised by the bin's mean RMV \cite{levi2022evaluating}. The chosen binning strategy can vary, from simpler options such as equal-size and equal-width bins, to more adaptive schemes that attempt to resolve the scaling of the ENCE with the square root of the number of bins $M^{1/2}$ for perfectly or almost perfectly calibrated models, as shown by \cite{pernot2023properties}. In addition, Pernot has shown that ENCE is not resistant to the effect of outliers \cite{pernot2023properties}.

\subsubsection{Bivariate histogram visualisation}

Plotting uncertainty vs.\ prediction error in a bivariate histogram can provide insight into the distribution of model outputs, providing the smallest-scale assessment of local calibration. This could allow the prevalence of outliers and how the model's predicted uncertainties capture the distributions of prediction errors to be assessed.

\subsection{Issues around metrics based on binning}\label{other}

\subsubsection{Binning strategies}
The method used to bin model predictions will have an affect on the values of the metrics. For example, \cite{pernot2023properties} mentions that the value of the ENCE (introduced in Section \ref{sec:ence}) can scale with the number of bins chosen. For now we advise using the binning strategies described in each metric's original paper. Metrics like the smECE, introduced in Section \ref{sec:smECE}, attempt to mitigate the issues surrounding binning strategies, through either kernel density estimation or determining the quality of calibration non-locally.  

\subsection{Comparing uncertainty reliability across different output types}

The outputs from the different UQ techniques are not consistent -- our implementation of Deep Ensembles and Monte Carlo Dropout output distributions that parametrise a Gaussian, from which the mean and standard deviation can be used to calculate regression metrics such as the ENCE or CRPS. In contrast, model results from conformal prediction are given as distribution-free coverage intervals. Evaluating the coverage between the different model outputs is trivial, as we can compute the quantiles from distributions using the predicted CDF, however if we wish to compare metric values such as the ENCE, more thought is required on how to convert between the output types. 

There exist multiple methods to convert between coverage intervals (which are either prediction sets or prediction intervals) to either probabilities for classification, or distributions for regression. These are briefly described in the next sections.

\subsubsection{Classification conversion methods} 

To compare the outputs from Venn ABERS conformal prediction to other UQ techniques, we must determine the best probability value from the intervals given about the model's prediction for each class. Choosing an optimal probabilistic prediction $p$ from the interval requires minimising our ``regret", which in our case is the log loss optimisation function. The solution is equivalent to taking the Jaccard mean of the two probabilities \cite{vovk2012venn}, defined as:
\begin{equation}
    p = \frac{p_{1}}{1-p_{0}+p_{1}},
\end{equation}
where $p$ is the class probability, and $(p_{0}, p_{1})$ are the lower and upper bounds of the interval. This allows us to calculate confidence based calibration metrics. Given we consider binary classification, this allows us to take the entropy of the distribution as well. 

\subsubsection{Regression conversion methods}
\label{sec:output_conversion}

For the case of evaluating the PICP and CCE, we need to obtain confidence intervals for each prediction from the output distribution. This can be calculated by determining the quantiles from the predicted CDF for each chosen confidence level. For UQ techniques that output distributions (DE and MCD) this is accomplished by finding the relevant quantiles for the Gaussian distribution parametrised by the model prediction and uncertainty.

\subsubsection{Prediction intervals to distributions}\label{sec:interval_conversion}

The outputs of conformal prediction for regression tasks are given as prediction intervals for a specified confidence level. The underlying distribution of these intervals is unknown, and indeed this distribution-free quality is an appealing attribute of the conformal prediction method. However, if we wish to evaluate metrics such as the ENCE that require information such as predicted mean and variance, we need to convert these intervals to distributions. 

The easiest method to do this would be to assume a Gaussian and, for a given confidence level, use the quantiles to obtain the distribution parameters. These can then be used as the model's predicted mean and variance to determine distribution-based metrics e.g.\ ENCE.

An alternative approach would be to randomly sample between the upper and lower bounds of the interval, and perform a kernel density estimate (KDE) over these. This estimated distribution can then be used to calculate the relevant distribution-based metrics, whilst the standard PICP metric can be determined from the intervals themselves. The motivation for this approach is to determine whether an alternative to the Gaussian assumption may yield better calibrated model uncertainties. The predicted standard deviation from mean-variance predictors only gives calibrated intervals if the residuals follow a Gaussian distribution, and this assumption does not always hold \cite{dewolf2023valid}. However, it is not the aim of this paper to develop a rigorous conversion method between distribution-free coverage intervals, and parametric distribution results. 

\section{Training details} 
\label{sec:train_details}
\subsection{Training details for AF classification}
\label{sec:train_details_af_tasks}
\begin{table}[h]
\centering
\begin{tabular}{|c|c|c|c|c|c|} 
\hline
\textbf{Task} & \textbf{Loss} & \textbf{Batch size} & \textbf{Learning rate} & \textbf{Weight decay}& \textbf{Optimiser}\\ \hline
AF alexnet & Algorithm 2 & 64 & 1e-5 & 1e-3& AdamW\\ \hline
AF resnet & Algorithm 2 & 64 &1e-4 & 1e-4 & AdamW\\ \hline
\end{tabular}
\caption{Training details for the AF classification task.}
\label{tab:train_details_af_tasks}
\end{table}

For MCD, we use a Dropout rate of 5\% for \textbf{resnet} and 20\% for \textbf{alexnet}. For DE we train 5 ensembles, where for \textbf{alexnet} we use Kaiming uniform initialisation and for \textbf{resnet}, we employ Kaiming normal initialisation for the convolutional and linear layers. 

We use an input length of 800 elements. The models have four output nodes, where two predict the logit for each class, and the other two predict the corresponding variance of the random noise used to corrupt each logit.

\subsubsection{Practical Implementation}
We train the models with the AdamW optimiser \cite{loshchilov2017decoupled} (which we choose instead of Adam, as Adam does not implement generic L2 regularisation), and likelihood-based losses to implement MAP estimation. We use a patience of 15 epochs and monitor the validation AUC to determine the best model. We use a plateau-based learning rate scheduler on the validation AUC, with a patience of 8 epochs and a multiplicative factor of 0.5.

\subsection{Training details for BP estimation}
\begin{table}[h]

\centering
\begin{tabular}{|c|c|c|c|c|c|} 
\hline
\textbf{Task} & \textbf{Loss} & \textbf{Batch size} & \textbf{Learning rate} & \textbf{Weight decay}& \textbf{Optimiser}\\ \hline
Calib/calibfree BP alexnet (not QR) & Equation (\ref{eq:L_BP}) & 32 & 1e-5 & 1e-3 & AdamW \\ \hline
Calib/calibfree BP resnet (not QR) & Equation (\ref{eq:L_BP}) & 32& 1e-4 & 1e-3 & AdamW\\ \hline
Calib/calibfree BP alexnet (QR) & Equation (\ref{eq:quantile}) & 32& 1e-5 & 1e-3 & AdamW\\ \hline
Calib/calibfree BP resnet (QR) & Equation (\ref{eq:quantile}) & 32& 1e-4 & 1e-3 & AdamW\\ \hline
\end{tabular}
\caption{Training details for the BP Regression task.}
\label{tab:train_details_bp_tasks}
\end{table}

For all non-QR models that predict the parameters of two Gaussian distributions, one for SBP and one for DBP, with means $\mu_{\text{SBP}}$ and $\mu_{\text{DBP}}$ and standard deviations $\sigma_{\text{SBP}}$ and $\sigma_{\text{DBP}}$, the loss for a single example is given by (\ref{eq:L_BP}):

\begin{equation}
    L_{BP} = \log\frac{\sigma_{\text{DBP}}}{2} + \frac{{(\mu_{\text{DBP}} - y_{\text{DBP}})}^2}{2\sigma_{\text{DBP}}^2} + \log\frac{\sigma_{\text{SBP}}}{2} + \frac{{(\mu_{\text{SBP}} - y_{\text{SBP}})}^2}{2\sigma_{\text{SBP}}^2}.
    \label{eq:L_BP}
\end{equation}
The value is aggregated with a mean over all the outputs for a given batch.

We use the same weight initialisation scheme and dropout rates as implemented for AF detection (see Appendix \ref{sec:train_details_af_tasks}). We use an input length of 1250 elements. Early stopping is employed based on the validation loss with a patience of 15 epochs. We also train these models with the AdamW optimiser \cite{loshchilov2017decoupled}.

For quantile regression, we use 10 output nodes, where 5 correspond to SBP and DBP quantiles respectively (prediction given by $\hat{y}_{\text{SBP}}$ and $\hat{y}_{\text{DBP}}$) and optimise the model with a quantile loss:
\begin{equation}
\label{eq:quantile}
L_{\text{QR}}=\frac{1}{10}\sum_{i=1}^5 \Big[ \text{max}[\nu_i(y_{\text{SBP}} - \hat{y}_{\text{SBP}}), (\nu_i-1)(y_{\text{SBP}} - \hat{y}_{\text{SBP}})] + \text{max}[\nu_i(y_{\text{DBP}} - \hat{y}_{\text{DBP}}), (\nu_i-1)(y_{\text{DBP}} - \hat{y}_{\text{DBP}})] \Big]
\end{equation}
where $y_{\text{SBP}}$ and $y_{\text{DBP}}$ are the ground truths, and $\pmb{\nu} = [0.0228, 0.1587, 0.5, 0.8413, 0.9772]$ are the quantile levels. As for the non-QR case, the value is aggregated with a mean over all the outputs for a given batch.

\section{AF classification predictive performance analysis}\label{sec:appendix_pred_performance}

This section presents a more detailed analysis of the predictive performance results for each UQ method addressed in Section \ref{sec:global}, using the binary classification metrics and associated thresholds outlined in \cite{moulaeifard2025machine}. Here, we quote the AUC, F1 score, Specificity (Spec), Sensitivity (Sens), and Matthew's Correlation Coefficient (MCC) for two thresholds that yield a Spec/Sens greater than $0.8$.  

\begin{table}[!ht]
    \centering
    \footnotesize
    \begin{tabular}{|c|*{5}{|c}|c|}
    \hline
        \multirow{2}{*}{\textbf{UQ type}} & \multicolumn{6}{c|}{\textbf{Predictive Performance Metrics}} \\ \cline{2-7}
        & \textbf{AUC↑} & \textbf{F1 (0.5)↑} & \textbf{Spec (Sens > 0.8)↑} & \textbf{Sens (Spec > 0.8)↑} & \textbf{MCC (Sens > 0.8)↑} & \textbf{MCC (Spec > 0.8)↑} \\ \hline\hline
        \textbf{MAP} & 0.81 & \textbf{0.66} & 0.66 & 0.63 & 0.45 & 0.44 \\ \hline
        \textbf{MCD} & 0.82 & 0.61 & 0.68 & 0.64 & 0.47 & 0.44 \\ \hline
        \textbf{DE} & \textbf{0.83} & 0.62 & \textbf{0.70} & \textbf{0.69} & \textbf{0.49} & \textbf{0.49} \\ \hline
        \textbf{MCD+TS} & 0.82 & 0.61 & 0.68 & 0.64 & 0.47 & 0.44 \\ \hline
        \textbf{MCD+IR} & 0.82 & 0.50 & 0.65 & 0.43 & 0.46 & 0.41 \\ \hline
        \textbf{DE+TS} & \textbf{0.83} & 0.62 & \textbf{0.70} & \textbf{0.69} & \textbf{0.49} & \textbf{0.49} \\ \hline
        \textbf{DE+IR} & \textbf{0.83} & 0.51 & 0.64 & \textbf{0.69} & 0.47 & \textbf{0.49} \\ \hline
        \textbf{Venn-ABERS} & 0.81 & 0.57 & 0.63 & 0.51 & 0.45 & 0.43 \\ \hline
    \end{tabular}
    \caption{\textbf{alexnet} predictive performance results for AF classification for each of the main UQ techniques. The best result is in bold, with the best result across both models underlined in addition.}
    \label{table:appendix_alexnet_predictive_results}
\end{table}

\begin{table}[!ht]
    \centering
    \footnotesize
    \begin{tabular}{|c|*{5}{|c}|c|}
    \hline
        \multirow{2}{*}{\textbf{UQ type}} & \multicolumn{6}{c|}{\textbf{Predictive Performance Metrics}} \\ \cline{2-7}
        & \textbf{AUC↑} & \textbf{F1 (0.5)↑} & \textbf{Spec (Sens > 0.8)↑} & \textbf{Sens (Spec > 0.8)↑} & \textbf{MCC (Sens > 0.8)↑} & \textbf{MCC (Spec > 0.8)↑} \\ \hline\hline
        \textbf{MAP} & 0.84 & 0.68 & 0.74 & 0.73 & 0.52 & 0.53 \\ \hline
        \textbf{MCD} & 0.85 & \underline{\textbf{0.70}} & \underline{\textbf{0.75}} & \underline{\textbf{0.75}} & \underline{\textbf{0.54}} & 0.54 \\ \hline
        \textbf{DE} & \underline{\textbf{0.86}} & 0.68 & \underline{\textbf{0.75}} & \underline{\textbf{0.75}} & \underline{\textbf{0.54}} & \underline{\textbf{0.55}} \\ \hline
        \textbf{MCD+TS} & 0.85 & \underline{\textbf{0.70}} & \underline{\textbf{0.75}} & \underline{\textbf{0.75}} & \underline{\textbf{0.54}} & 0.54 \\ \hline
        \textbf{MCD+IR} & 0.85 & 0.60 & 0.74 & 0.60 & 0.53 & 0.51 \\ \hline
        \textbf{DE+TS} & \underline{\textbf{0.86}} & 0.68 & \underline{\textbf{0.75}} & \underline{\textbf{0.75}} & \underline{\textbf{0.54}} & \underline{\textbf{0.55}} \\ \hline
        \textbf{DE+IR} & 0.85 & 0.59 & 0.71 & 0.74 & 0.53 & \underline{\textbf{0.55}} \\ \hline
        \textbf{Venn-ABERS} & 0.84 & 0.63 & 0.73 & 0.71 & 0.52 & 0.52 \\ \hline
    \end{tabular}
    \caption{\textbf{resnet} predictive performance results for AF classification for each of the main UQ techniques. The best result is in bold, with the best result across both models underlined in addition.}
    \label{table:appendix_resnet_predictive_results}
\end{table}

\bibliographystyle{unsrt}  
\bibliography{references}

\end{document}